\documentclass[letterpaper]{fairmeta}

\usepackage{amsmath,amssymb,pifont,tabularx}

\AtBeginDocument{\let\cite\citep}

\let\setpaperabstract\abstract
\RenewDocumentEnvironment{abstract}{+b}{\gdef\paperabstract{#1}}{}

\tcbuselibrary{skins,breakable,listings}

\newtcblisting{stvprompt}[1]{%
  enhanced,
  breakable,
  listing only,
  listing engine=listings,
  listing options={
    basicstyle=\ttfamily\scriptsize,
    breaklines=true,
    breakatwhitespace=false,
    columns=fullflexible,
    keepspaces=true,
    showstringspaces=false
  },
  colback=gray!3,
  colframe=gray!60,
  boxrule=0.35pt,
  arc=1mm,
  title={#1},
  fonttitle=\bfseries\small
}

\title{Generalist Meets Specialists: An Evolving Multimodal Agent with Persistent Tool Orchestration for Traffic Scenario Understanding}
\author[1,2,\dagger]{Linjia Kang}
\author[1,\dagger]{Binhan Jiang}
\author[1]{Juan Yang}
\author[1]{Zhiqin Song}
\author[1,3]{Xiangsen Wu}
\author[4]{Qionglin Ren}
\author[1]{Jiang Liu}
\author[1]{Zhongjiang He}
\author[1,*]{Hao Sun}

\affiliation[1]{Xingchen AGI Lab, China Telecom Artificial Intelligence Technology (Beijing) Co., Ltd.}
\affiliation[2]{Tsinghua University}
\affiliation[3]{Karlsruher Institut f\"ur Technologie}
\affiliation[4]{Fudan University}
\contribution[\dagger]{Equal contribution.}
\contribution[*]{Corresponding author.}

\begin{document}
\begin{abstract}
Traffic scenario understanding requires combining
fine-grained visual evidence with spatial and temporal
reasoning. Tool-augmented vision-language models
(VLMs) can use specialized perception tools, but
effective collaboration faces two challenges:
terminal rewards provide limited guidance for
intermediate decisions and error recovery, while
repeated planning adds unnecessary cost across
similar tasks.
Echoing the fast--slow dichotomy of human cognition,  
we propose \textbf{StarTV-Agent},
which combines policy learning with experience
evolution to support adaptive reasoning when
needed and efficient reuse of prior experience
when possible.
During training, Role-Sensitive Process Optimization
(RSPO) combines process advantages with terminal
advantages, while Weakness-Driven Demonstration
Learning (WDL) provides validated teacher
demonstrations targeting the student's weak questions.
During inference, Lightweight Adaptive Workflow Reuse
(LAWR) enables a fast path by selecting suitable
workflows and binding task-specific arguments and
resource references, while a slow ReAct-based path
handles tasks without suitable workflows.
We further design label-free feedback based on
execution completion, workflow adoption, and cost
savings to guide workflow reuse and memory updates.
Across three challenging benchmarks, StarTV-Agent achieves up to 8.0\% relative performance improvement, while reducing token consumption by up to 52.1\%.
\end{abstract}

\setpaperabstract{\paperabstract}
\maketitle

\section{Introduction}
\label{sec:intro}

Determining whether a vehicle ran a red light may seem
simple, but it requires several linked visual judgments:
identifying the vehicle and traffic signal, locating
the stop line, tracking the vehicle, and checking
whether it crossed while the signal was red.
An error in any step can change the final answer.
Recent studies~\cite{zhou2025tumtraf,parikh2025roadsocial,
guan2026roadscenevqa,wei2025driveqa,ma2026drivecombo}
show that traffic understanding requires both
fine-grained perception and reasoning over spatial
relationships, temporal events, and traffic rules.
Beyond recognizing objects, an agent must decide
what evidence to collect, which tools to use,
and whether the evidence supports its answer.

Tool-augmented vision-language models (VLMs) combine
general reasoning with specialized perception
tools~\cite{zhang2026vital,lu2026vistagym,
chen2026videochatm1,ke2025dwim}.
Recent work such as TAU-Agent~\cite{lin2026tauagent}
has brought this approach to traffic understanding.
However, using expert tools introduces a series
of linked decisions, from planning and expert selection
to argument generation and evidence checking.
The agent must also recognize failed tool calls
and correct earlier decisions when needed.
Learning these behaviors is difficult with terminal
rewards alone, which provide little guidance on where
an interaction succeeds or fails.
Recent studies explore finer-grained feedback for
agent training~\cite{cheng2026graphgpo,wang2026beacon,
zhang2026qevolve}, but effective tool collaboration
also calls for supervision that reflects each
decision's role. Teacher demonstrations can provide
successful examples~\cite{li2026guidedopd}, yet
their value depends on whether they address
the student's actual weaknesses.

More importantly, real-world traffic monitoring systems demand timely
responses, while many traffic questions share
reusable tool-use patterns. However, their target
objects, arguments, and evidence dependencies
may differ.
For example, two red-light questions may follow
the same detection and tracking steps but involve
different vehicles and traffic signals.
Reusing a workflow therefore requires adapting
its inputs rather than copying past arguments.
Moreover, the agent must decide which workflows
are worth reusing and updating.
Recent work on agent memory and experience
~\cite{jiang2026xskill,feng2026expweaver,
cai2026memopilot} motivates learning from past
interactions, but efficient reuse also requires
simple selection and feedback from actual execution.
These challenges suggest a natural division:
a slow reasoning path for tasks requiring adaptive
tool planning, and a fast execution path for tasks
with reusable tool-use patterns.

To address these challenges, we propose
\textbf{StarTV-Agent}, which combines
\emph{policy learning} and \emph{experience evolution}
for reliable and efficient generalist--specialist
collaboration. Built on a shared VLM policy and
frozen expert tools, StarTV-Agent supports
complementary slow and fast paths: the former
uses ReAct-based orchestration for adaptive
reasoning, while the latter directly executes
suitable parameterized workflows.
Execution feedback further connects the two paths
by turning useful tool-use patterns into
reusable experience.

For \emph{policy learning}, we introduce
Role-Sensitive Process Optimization (RSPO)
and Weakness-Driven Demonstration Learning (WDL).
RSPO combines terminal advantages with process
advantages that compare decisions by their roles
in the interaction, providing feedback for tool
use and error recovery.
WDL identifies questions the student struggles
with and uses an independent teacher to generate
successful tool interactions. Demonstration trajectories
are then used to fine-tune the trained policy.
Together, RSPO and WDL improve learning from
both intermediate feedback and complete examples,
strengthening the agent's slow reasoning path.

For \emph{experience evolution}, we propose
Lightweight Adaptive Workflow Reuse (LAWR).
It enables the fast path through lightweight
hybrid search and parameterized workflow execution.
The selected workflow keeps its tool sequence
but sets new arguments for the current task.
When no suitable workflow exists, the agent
uses the slow ReAct-based reasoning path.
After execution, the agent updates its workflow
memory using our label-free feedback signals
based on execution completion, workflow adoption,
and cost savings.
This allows useful execution patterns to be
reused and refined over time, reducing the need
for repeated planning.

Our main contributions are summarized as follows:
\begin{itemize}
    \item We introduce RSPO and WDL, combining
    role-sensitive process advantages with
    targeted teacher demonstrations to improve
    multi-step tool orchestration.

    \item We propose LAWR, connecting slow
    reasoning with fast workflow execution
    through task-adaptive reuse and label-free
    experience evolution.

    \item We conduct extensive experiments on
    three traffic scenario understanding benchmarks,
    validating the effectiveness and efficiency
    of StarTV-Agent.
\end{itemize}

\section{Related Work}
\label{sec:related_work}

\subsection{Traffic Scenario Understanding}
\label{sec:related_traffic_agents}

Traffic scenario understanding requires more than detecting
road objects. It also involves reasoning about their
spatial relationships, motion, and traffic rules.
Driving benchmarks study scene reasoning, rare events,
and traffic-rule knowledge~\cite{sima2024drivelm,
chen2025codalm,wei2025driveqa,ma2026drivecombo},
while roadside benchmarks focus on traffic events
and interactions among road users~\cite{zhou2025tumtraf,
parikh2025roadsocial,guan2026roadscenevqa}.
Recent multimodal benchmarks further include audio
and thermal inputs~\cite{xing2025echotraffic,
zhang2026trafficvqa}.
Together, these tasks require agents to combine
evidence across objects, frames, and input types
rather than relying on single visual observations.

Following ReAct~\cite{yao2023react}, visual agents
combine reasoning with tool use to solve complex
tasks. General frameworks such as
OctoTools~\cite{lu2026octotools} support planning
and execution across different tools.
Other studies improve visual evidence collection
through frame selection and temporal
navigation~\cite{zhang2026vital,li2026lenswalk,
qiu2026longvideor1,tan2026msjoe,tang2026tspo}.
Recent methods also explore multi-step tool use,
workflow generation, and RL for visual reasoning~\cite{yin2025toolvqa,
ke2025dwim,lu2026vistagym,guo2026codevision,
chen2026videochatm1,li2026agentflow}.
Structured visual representations provide another
way to support reasoning~\cite{janjua2026perceptionprograms,
li2026geoanchor,gao2026motto},
while TAU-Agent~\cite{lin2026tauagent} applies
tool-based collaboration to traffic anomaly
understanding.
These studies show the value of specialized tools,
but effective collaboration also requires knowing
when to reason carefully through tool interactions
and when familiar tool-use patterns can be reused
without repeated planning.

\subsection{Policy Learning and Experience Evolution}
\label{sec:related_learning_evolution}

\paragraph{Policy learning.}
Reinforcement Learning~(RL) has improved planning
and tool use in visual agents~\cite{li2026agentflow,
zhang2026vital}. However, rewards based on final
outcomes provide limited guidance for individual
decisions in multi-step interactions.
Beyond sequence-level methods such as Group Sequence
Policy Optimization (GSPO)~\cite{zheng2025gspo}, recent studies improve
intermediate feedback through state grouping,
transition modeling, and milestone-based
rewards~\cite{feng2025gigpo,cheng2026graphgpo,
wang2026beacon}. Others explore process supervision,
step-level optimization, and tool-based reward
models~\cite{guo2026codevision,chen2026videochatm1,
zhang2026qevolve,liu2026mlagent,ding2026armthinker,gao2026sapo}.
Teacher guidance and experience-based learning
provide additional ways to train agents
~\cite{zhang2026earlyexperience,li2026guidedopd,huang2026calibrl}.
In generalist--specialist collaboration, errors in
expert selection, tool arguments, or evidence checking
can lead to the same wrong answer, making it difficult
to identify which decisions need improvement.
Moreover, process feedback alone may not teach the
agent how to solve tasks beyond its current ability.
To strengthen the slow reasoning path, StarTV-Agent
combines RSPO for role-sensitive process feedback
with WDL for targeted demonstrations on difficult tasks.

\paragraph{Experience evolution.}
Prior work shows that agents can improve their decisions
by learning from past interactions~\cite{shinn2023reflexion,
zhao2024expel,wu2026evolver}. Recent methods retrieve and
update stored experience~\cite{feng2026expweaver,cai2026memopilot}
or turn past executions into reusable
procedures~\cite{mi2026skillpro,shao2026nsi,wei2026evoc2f}.
FlowEvo~\cite{ren2026flowevo} distills successful workflows
into skills that can be executed directly or used to guide
later planning. Multimodal agents also learn reusable tools
and skills from visual experience~\cite{huang2026meta,jiang2026xskill}.
In traffic scenario understanding, similar questions can
concern different road users or require different visual
evidence. A past workflow may therefore miss evidence
needed for the current question, while planning from
scratch for every question adds unnecessary cost.
LAWR addresses this trade-off by reusing workflows that
fit the current task and using adaptive planning for the
remaining tasks. Execution feedback guides which workflows
to reuse or update over time.

\begin{figure*}[t]
    \centering
    \includegraphics[width=1.0\linewidth]{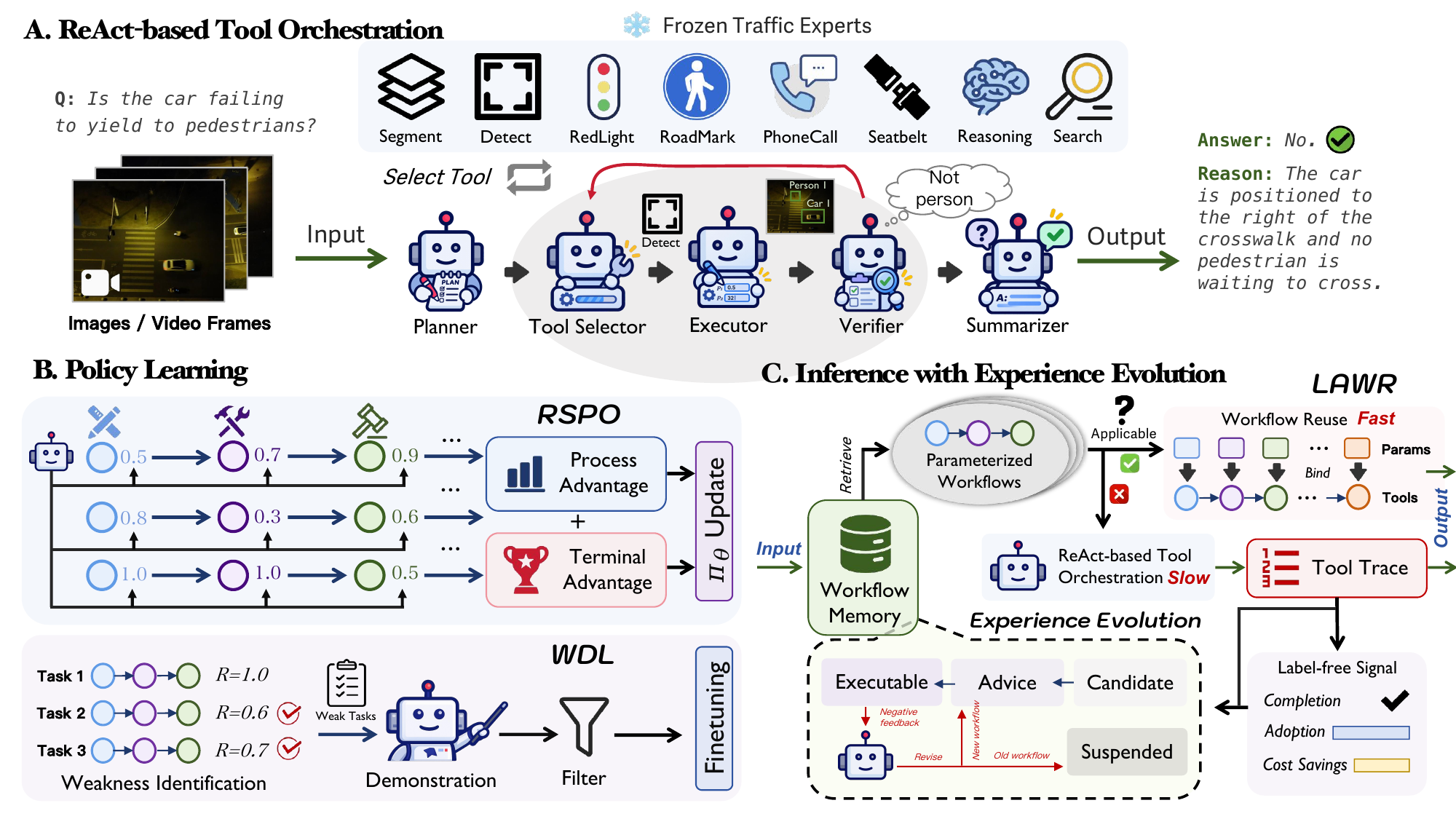}   
    \caption{\textbf{Overview of StarTV-Agent.}
    \textbf{(A) ReAct-based Tool Orchestration:}
    A shared VLM policy coordinates frozen traffic experts
    through task planning, tool selection, argument
    generation, and evidence verification.
    \textbf{(B) Policy Learning:}
    RSPO combines role-sensitive process and terminal
    advantages, while WDL provides targeted teacher
    demonstrations for weak tasks.
    \textbf{(C) Inference with Experience Evolution:}
    The agent follows a fast path by reusing suitable
    parameterized workflows or a slow ReAct-based path
    for adaptive reasoning. Label-free execution feedback
    guides workflow reuse and memory updates.}
    \label{fig:framework}
\end{figure*}

\section{Methodology}
\label{sec:method}

\subsection{Framework Overview}
\label{sec:method_overview}

Given a traffic input $x$ (images or video frames)
and a question $q$, StarTV-Agent uses a vision-language
policy $\pi_\theta$ to coordinate frozen domain-specific
experts $\mathcal{E}$, including detection, segmentation,
and attribute recognition
(Appendix~\ref{app:tool_inventory}).
As shown in Fig.~\ref{fig:framework}, the framework
combines ReAct-based tool orchestration,
policy learning, and experience evolution to
support complementary slow and fast execution paths.
Detailed prompts for StarTV-Agent are provided in
Appendix~\ref{app:training_prompts}.

\noindent\textbf{(A) ReAct-based Tool Orchestration.}
The slow path uses ReAct-based tool
orchestration~\cite{yao2023react} for adaptive
planning and error recovery.
The policy makes decisions based on the current
state and decision role:
\begin{equation}
y_t \sim \pi_\theta(\cdot \mid s_t,d_t),
\qquad
s_t=(q,x,H_t,\mathcal{R}_t,b_t),
\label{eq:react_policy}
\end{equation}
where $H_t$ is the interaction history, $\mathcal{R}_t$
stores the collected evidence, and $b_t$ is the
remaining budget. 
The role $d_t$ takes five values:
Planner (initial task planning),
Tool Selector (expert tool selection and sub-goal decision),
Executor (tool argument and resource-reference generation),
Verifier (continuation decision), and
Summary (evidence-based answering).
Tool results are stored as typed resources with
unique identifiers and source dependencies.
After each tool call, the Verifier checks whether
the evidence is sufficient and whether further
planning or error recovery is needed.

\noindent\textbf{(B) Policy Learning.}
During training, 
RSPO combines role-sensitive process advantages with terminal advantages to guide intermediate
decisions, while WDL provides validated teacher
demonstrations targeting the student's weaknesses.
Together, they strengthen multi-step tool orchestration
and error recovery.

\noindent\textbf{(C) Inference with Experience Evolution.}
During inference, the agent first checks for
a reusable workflow.
If a suitable workflow is available, the fast
path executes its stored tool sequence with
task-specific arguments and resource references,
avoiding repeated planning.
Otherwise, the agent uses the slow ReAct-based path.
After submitting the final answer, the agent
updates its external workflow memory using
label-free feedback based on execution completion,
workflow adoption, and cost savings.
This allows useful tool-use patterns to be
reused and refined for future tasks without
changing the VLM policy.

\subsection{Policy Learning for StarTV-Agent}
\label{sec:method_policy}

The slow reasoning path supports adaptive tool
planning and error recovery, but terminal rewards
provide limited feedback on intermediate decisions.
StarTV-Agent therefore combines RSPO and WDL
to improve multi-step tool orchestration.

\subsubsection{Role-Sensitive Process Optimization}
For each training question, multiple student trajectories
are sampled with task-level rewards $R_i$.
A process evaluator assigns a score $p_{i,t}\in[0,1]$
to each decision based on its role, available evidence,
and tool feedback. These scores cover all five roles,
including error correction after failed tool calls.
The evaluator may access training references but
returns only scalar scores to the student.

RSPO combines terminal and process advantages:
\begin{equation}
\begin{aligned}
A_i^{\mathrm{term}}
&=
\frac{R_i-\mu_q^{\mathrm{term}}}
{\sigma_q^{\mathrm{term}}+\epsilon},\\
A_{i,t}^{\mathrm{proc}}
&=
\frac{p_{i,t}-\mu_{q,h}^{\mathrm{proc}}}
{\sigma_{q,h}^{\mathrm{proc}}+\epsilon},
\end{aligned}
\label{eq:process_advantages}
\end{equation}
where $h$ denotes the decision role and its occurrence
index, and $\mu$ and $\sigma$ are the corresponding
group statistics. The terminal advantage compares
trajectories for the same question, while the process
advantage compares decisions with the same question,
role, and occurrence index. They are combined as
\begin{equation}
A_{i,t}
=
\alpha A_i^{\mathrm{term}}
+
\beta A_{i,t}^{\mathrm{proc}},
\qquad \alpha+\beta=1.
\label{eq:combined_advantage}
\end{equation}

The student is trained with a GSPO-style
length-normalized, sequence-level clipped
objective~\cite{zheng2025gspo}, using $A_{i,t}$
for each role-conditioned response.

\subsubsection{Weakness-Driven Demonstration Learning}
Process feedback alone may not show the student
how to solve tasks beyond its current ability.
WDL therefore adds teacher demonstrations
for questions the student still struggles with.

\noindent\textbf{Weakness identification.}
After RL training, student performance on each
question is measured by the mean raw task reward:
\begin{equation}
\bar{R}_{q}
=
\frac{1}{|\mathcal{G}_{q}|}
\sum_{R \in \mathcal{G}_{q}} R,
\label{eq:weakness_score}
\end{equation}
where $\mathcal{G}_{q}$ contains deduplicated,
scorable rewards from valid trajectories.
Questions with $\bar{R}_{q}<\eta$ are selected,
where $\eta=0.8$.

\noindent\textbf{Independent demonstration generation.}
An independent teacher solves each selected question
from scratch using the same tools and execution budget,
without access to student trajectories, rewards,
or reference answers. A separate filter retains
demonstrations that complete successfully, pass
evidence-source and output-format checks, and meet
task-specific reward requirements: exact correctness
for binary, multiple-choice, and short-answer questions,
or a reward of at least $0.8$ otherwise.

\noindent\textbf{Targeted supervised fine-tuning.}
Accepted model-call events form the training set
$\mathcal{D}$, preserving their decision contexts
and action formats. All five roles are trained
with the Supervised Fine-Tuning~(SFT) loss:
\begin{equation}
\mathcal{L}_{\mathrm{SFT}}(\theta)
=
-\sum_{k\in\mathcal{D}}
\sum_{t=1}^{T_k}
\log \pi_{\theta}
\left(
a_{k,t}\mid x_k,a_{k,<t}
\right),
\label{eq:demo_sft}
\end{equation}
where $x_k$ is the decision context and $a_{k,1:T_k}$
are the target action tokens. Prompt and image tokens
are excluded from the loss.
Fine-tuning the RL-trained policy on these
demonstrations provides complete examples of
successful tool orchestration, strengthening
the shared policy for deliberate reasoning.

\subsection{Experience Evolution for StarTV-Agent}
\label{sec:experience_evolution}

Similar traffic questions often share tool-use
patterns but differ in target objects, arguments,
and resource dependencies.
Repeated planning incurs extra cost, while
replaying fixed trajectories may fail on new inputs.
To support efficient reuse, StarTV-Agent maintains
an external memory $\mathcal{M}^{(n)}$ of
parameterized workflows:
\begin{equation}
\mathcal{M}^{(n)}
=
\left\{
e_j=(W_j,g_j,z_j,\mathbf{s}_j)
\right\}_j,
\label{eq:experience_memory}
\end{equation}
where $W_j$ defines a tool sequence and a
directed acyclic graph (DAG) of resource
dependencies. Each node represents a typed
resource placeholder, and each edge specifies
a dependency.
Here, $g_j$ specifies applicability conditions,
$z_j$ denotes the workflow state, and
$\mathbf{s}_j$ stores usage statistics.
The index $n$ denotes successive tasks.

LAWR consists of four stages: workflow retrieval,
parameterized execution, label-free feedback,
and experience evolution.

\subsubsection*{Stage 1: Workflow Retrieval and Selection}
LAWR ranks stored workflows by combining keyword and
semantic search with reciprocal rank fusion. It then
checks each workflow's applicability conditions against
the current question and input metadata without additional
VLM calls. LAWR selects the highest-ranked executable
workflow whose conditions are met. If no workflow
qualifies, the agent follows the slow ReAct path.

\subsubsection*{Stage 2: Parameterized Workflow Execution}
During execution, LAWR adapts the selected workflow $W^\star$
to the current task through parameter generation.
Tool arguments $p_i$ and resource selectors
$\rho_i$ are generated based on the question,
input metadata, and tool requirements.
At each step $i$, the DAG guides resource
resolution from $\mathcal{R}_{i-1}$.
Tool outputs are registered in $\mathcal{R}_i$
with their source dependencies.
This enables the reuse of established tool-use
patterns while adapting arguments and resource
references to new inputs, avoiding the limitations
of replaying fixed execution trajectories.

\subsubsection*{Stage 3: Label-Free Execution Feedback}
After the final answer is submitted, LAWR
evaluates workflow use through execution
completion, tool adoption, and cost savings.
The latter two are defined as
\begin{equation}
\begin{aligned}
\operatorname{Adopt}(e,\tau)
&=
\frac{\operatorname{Match}(T(e),T(\tau))}
{|T(e)|},\\
S(\tau)
&=
\frac{C_{\mathrm{ref}}-C(\tau)}
{C_{\mathrm{ref}}}.
\end{aligned}
\label{eq:workflow_utility}
\end{equation}
Here, $T(e)$ and $T(\tau)$ denote the
prescribed and executed tool sequences,
respectively.
$\operatorname{Match}$ counts their shared
tool calls, including repeated calls.
$S(\tau)$ measures relative cost savings,
where $C_{\mathrm{ref}}$ is updated using
an exponential moving average of completed
executions without workflow reuse.
All signals are computed from execution
traces without additional model calls.

\subsubsection*{Stage 4: Execution-Driven Experience Evolution}
To maintain reliable workflow reuse, LAWR
updates the external memory using execution
feedback:
\begin{equation}
\mathcal{M}^{(n+1)}
=
U\left(\mathcal{M}^{(n)},\tau_n\right),
\label{eq:experience_update}
\end{equation}
where $\tau_n$ records tool calls, execution
status, costs, and workflow usage.

Using the signals in Eq.~\eqref{eq:workflow_utility},
LAWR evaluates each workflow use through
execution completion, tool adoption, and
relative cost savings.
For workflow $e_j$, $\bar S_j$ averages the
latest $K_S$ cost observations and
remains unavailable until $K_S$ observations
are collected.
A use receives positive feedback when
\begin{equation}
\begin{aligned}
w_j(\tau) ={}& u(\tau)\land\bigl(
\operatorname{Adopt}(e_j,\tau)\ge a_{\min}\bigr)\\
&\land\bigl(\bar S_j>s_{\min}),
\end{aligned}
\label{eq:workflow_win}
\end{equation}
where $u(\tau)$ indicates execution completion.
The empirical positive feedback rate is
$\mathrm{WR}_j=\mathrm{wins}_j/\mathrm{uses}_j$,
where $\mathrm{uses}_j$ counts evaluated uses
and $\mathrm{wins}_j$ counts positive outcomes.

Each workflow maintains a state
$z_j \in$ \{\texttt{candidate}, \texttt{advice},
\texttt{executable}, \texttt{suspended}\},
which controls its reuse eligibility.
New workflows start as \texttt{candidate}
and advance to \texttt{advice} when
$\mathrm{uses}_j\ge N_{\min}$ and
$\mathrm{WR}_j\ge\gamma_{\mathrm{prom}}$.
An additional positive use promotes them to
\texttt{executable}, enabling direct reuse.
An invalid use immediately demotes an
\texttt{executable} workflow to \texttt{advice}.
This gradual promotion and immediate demotion
limit the reuse of workflows with
unreliable execution feedback.

Workflows with at least
$K_{\mathrm{rev}}$ consecutive invalid uses
are considered for revision.
Each revision creates a new version in
\texttt{advice} and suspends the previous one.
Completed executions not covered by existing
workflows may generate new \texttt{candidate}
workflows.
New and revised workflows must pass tool
argument and resource-dependency checks
before storage.
Memory updates affect only future tasks,
allowing reusable tool-use patterns to
improve execution efficiency over time.

\section{Experiments}
\label{sec:experiments}

\begin{table*}[t]
\centering
\caption{Performance comparison on TUMTraf VideoQA.
We report multiple-choice accuracy (\%).
Overall is the mean accuracy across ten groups.
N/D indicates undisclosed parameters.
Bold and underline indicate the best and second-best results, respectively, excluding frontier-scale VLMs.}
\label{tab:tumtraf_results}

\scriptsize
\setlength{\tabcolsep}{2.5pt}
\renewcommand{\arraystretch}{1.12}

\begin{tabular*}{0.98\textwidth}{
@{\extracolsep{\fill}}lcccccccccccc@{}}
\toprule
& &
\multicolumn{2}{c}{\textbf{Positioning}} &
\multicolumn{2}{c}{\textbf{Counting}} &
\multicolumn{2}{c}{\textbf{Motion}} &
\multicolumn{2}{c}{\textbf{Class}} &
\multicolumn{2}{c}{\textbf{Existence}} &
\multirow{2}{*}{\textbf{Overall} $\uparrow$} \\

\cmidrule(lr){3-4}
\cmidrule(lr){5-6}
\cmidrule(lr){7-8}
\cmidrule(lr){9-10}
\cmidrule(lr){11-12}

\textbf{Method} & \textbf{Params}
& \textbf{E} & \textbf{H}
& \textbf{E} & \textbf{H}
& \textbf{E} & \textbf{H}
& \textbf{E} & \textbf{H}
& \textbf{E} & \textbf{H} & \\
\midrule

DeepSeek-V4.1-Flash & 552B
& 33.3 & 42.4 & 72.7 & 60.6 & 72.7 & 62.6
& 55.6 & 40.4 & 85.9 & 70.7 & 59.7 \\

Kimi K3 & 2.8T
& 40.4 & 40.4 & 82.8 & 65.7 & 68.7 & 54.5
& 60.6 & 53.5 & 91.9 & 69.7 & 62.8 \\

Qwen3.8-Max & 2.4T
& 55.6 & 38.4 & 80.8 & 66.7 & 71.7 & 75.8
& 73.7 & 74.7 & 90.9 & 72.7 & 70.1 \\

GPT-5.5 & N/D
& 49.5 & 41.4 & 76.8 & 63.6 & 69.7 & 71.7
& 65.7 & 50.5 & 86.9 & 69.7 & 64.5 \\

\midrule

Qwen3-VL-8B-Instruct & 8B
& 20.2 & \textbf{57.6}
& 68.7 & \textbf{65.7}
& 67.7 & 30.3
& 54.6 & 46.5
& \underline{87.9} & 59.6 & \underline{55.9} \\

STEP3-VL-10B & 10B
& 26.3 & 41.4
& 71.7 & 51.5
& \textbf{69.7} & 40.4
& \textbf{58.6} & 43.4
& 84.8 & 61.6 & 54.9 \\

Qwen3.8-27B & 27B
& 29.3 & 37.4
& \underline{76.8} & \underline{64.6}
& 64.6 & 27.3
& 57.6 & 41.4
& \textbf{91.9} & 57.6 & 54.8 \\

\midrule

DeepEyes-7B & 7B
& \underline{30.3} & \underline{53.5}
& 68.7 & 55.5
& 63.6 & 21.2
& \underline{58.5} & 45.4
& 82.8 & 56.5 & 53.6 \\

VTool-R1-7B & 7B
& 25.5 & 47.4
& 68.7 & 55.5
& 56.5 & 22.2
& 55.5 & 34.3
& 84.8 & 53.5 & 50.4 \\

TAU-Agent & 8B
& 22.2 & 38.3 & 63.6 & 59.5 & 44.4 & \textbf{71.7}
& 53.5 & \underline{55.5} & 78.7 & \underline{62.6} & 55.0 \\

\midrule

\textbf{StarTV-Agent} & 8B
& \textbf{38.4} & 36.4
& \textbf{83.8} & 54.5
& \underline{68.7} & \underline{41.4}
& 39.4 & \textbf{60.6}
& \underline{87.9} & \textbf{71.7}
& \textbf{58.3} \\

\bottomrule
\end{tabular*}
\end{table*}

\begin{table}[t]
\centering
\caption{Performance comparison on Traffic-VQA.
We report overall accuracy (\%) for each modality.
Mean is the arithmetic average of the three
modality-specific accuracy scores.
N/D indicates undisclosed parameters.
Bold and underline indicate the best and
second-best results, respectively, excluding
frontier-scale VLMs.}
\label{tab:traffic_results}

\scriptsize
\setlength{\tabcolsep}{1.8pt}
\renewcommand{\arraystretch}{1.12}

\begin{tabular*}{\columnwidth}{
@{\extracolsep{\fill}}
p{0.34\columnwidth}
c
cccc
@{}}
\toprule
\textbf{Method}
& \textbf{Params}
& \shortstack[c]{\textbf{OPT} $\uparrow$}
& \shortstack[c]{\textbf{TIR} $\uparrow$}
& \shortstack[c]{\textbf{MUL} $\uparrow$}
& \shortstack[c]{\textbf{Mean} $\uparrow$} \\
\midrule

DeepSeek-V4.1-Flash & 552B
& 40.1 & 40.1 & 46.2 & 42.1 \\

Kimi K3 & 2.8T
& 42.3 & 43.7 & 48.4 & 44.8 \\

Qwen3.8-Max & 2.4T
& 49.1 & 50.2 & 52.0 & 50.4 \\

GPT-5.5 & N/D
& 48.0 & 46.6 & 48.7 & 47.8 \\

\midrule

Qwen3-VL-8B-Instruct & 8B
& 42.3 & 40.9 & 43.7 & 42.3 \\

STEP3-VL-10B & 10B
& 36.2 & 36.9 & 37.6 & 36.9 \\

Qwen3.8-27B & 27B
& 40.5 & 25.8 & \textbf{49.8} & 38.7 \\

\midrule

DeepEyes-7B & 7B
& 43.3 & 36.9 & 37.6 & 39.3 \\

VTool-R1-7B & 7B
& 41.9 & 35.8 & 41.9 & 39.9 \\

TAU-Agent & 8B
& \underline{44.1} & \underline{41.9}
& 43.3 & \underline{43.1} \\

\midrule

\textbf{StarTV-Agent} & 8B
& \textbf{44.4} & \textbf{45.4}
& \underline{47.2} & \textbf{45.7} \\

\bottomrule
\end{tabular*}
\end{table}

\begin{table*}[t]
\centering
\caption{Performance comparison on RoadSceneVQA.
All metrics retain their native scales, with higher values
indicating better performance.
N/D indicates undisclosed parameters.
Bold and underline indicate the best and second-best results, respectively, excluding frontier-scale VLMs.}
\label{tab:roadscene_results}

\scriptsize
\setlength{\tabcolsep}{2.5pt}
\renewcommand{\arraystretch}{1.12}

\begin{tabular*}{0.98\textwidth}{
@{\extracolsep{\fill}}lcccccccc@{}}
\toprule
\textbf{Method} & \textbf{Params}
& \textbf{Exact Match} $\uparrow$
& \textbf{ROUGE-L} $\uparrow$
& \textbf{BLEU-4} $\uparrow$
& \textbf{METEOR} $\uparrow$
& \textbf{CIDEr} $\uparrow$
& \textbf{SPICE} $\uparrow$
& \textbf{GPT-Score} $\uparrow$ \\
\midrule

DeepSeek-V4.1-Flash & 552B
& 0.189 & 0.367 & 0.104 & 0.194
& 1.210 & 0.128 & 0.703 \\

Kimi K3 & 2.8T
& 0.155 & 0.348 & 0.068 & 0.202
& 1.016 & 0.111 & 0.706 \\

Qwen3.8-Max & 2.4T
& 0.237 & 0.392 & 0.061 & 0.212
& 1.045 & 0.095 & 0.811 \\

GPT-5.5 & N/D
& 0.145 & 0.355 & 0.104 & 0.208
& 1.135 & 0.122 & 0.752 \\

\midrule

Qwen3-VL-8B-Instruct & 8B
& \textbf{0.221} & \underline{0.359} & 0.077 & 0.168
& 1.105 & 0.097 & 0.643 \\

STEP3-VL-10B & 10B
& 0.111 & 0.302 & \underline{0.102} & 0.212
& 1.037 & 0.141 & 0.595 \\

Qwen3.8-27B & 27B
& \underline{0.168} & \textbf{0.368} & 0.099 & \underline{0.215}
& \underline{1.180} & 0.128 & \textbf{0.712} \\

\midrule

DeepEyes-7B & 7B
& 0.049 & 0.300 & \textbf{0.107} & \textbf{0.236}
& \textbf{1.277} & \textbf{0.170} & 0.599 \\

VTool-R1-7B & 7B
& 0.158 & 0.335 & 0.093 & 0.189
& 1.093 & 0.125 & 0.579 \\

TAU-Agent & 8B
& 0.032 & 0.238 & 0.054 & 0.205
& 0.702 & 0.141 & 0.637 \\

\midrule

\textbf{StarTV-Agent} & 8B
& 0.040 & 0.246 & 0.055 & 0.197
& 0.834 & \underline{0.149} & \underline{0.694} \\

\bottomrule
\end{tabular*}
\end{table*}

\begin{table}[t]
\centering
\caption{Ablation of policy learning components.
RSPO denotes role-sensitive process optimization.
Demo SFT and Weak Sel. denote demonstration
fine-tuning and weakness-based question selection.
The w/o Weak Sel. variant uses balanced random
sampling instead. Bold and underline indicate
the best and second-best results, respectively.}
\label{tab:training_ablation}

\scriptsize
\setlength{\tabcolsep}{1pt}
\renewcommand{\arraystretch}{1.12}

\begin{tabular*}{\columnwidth}{
@{\extracolsep{\fill}}lcccccc@{}}
\toprule

\textbf{Variant}
& \textbf{RL}
& \textbf{RSPO}
& \shortstack{\textbf{Demo}\\\textbf{SFT}}
& \shortstack{\textbf{Weak}\\\textbf{Sel.}}
& \shortstack{\textbf{RoadSceneVQA}\\
\textbf{GPT-Score} $\uparrow$}
& \shortstack{\textbf{Traffic-VQA}\\
\textbf{OA} $\uparrow$} \\

\midrule

\textit{Qwen3-VL-8B-Instruct}
& \ding{55} & \ding{55} & \ding{55} & --
& 0.620 & 41.3 \\

\midrule

\textit{Terminal-only RL}
& \ding{51} & \ding{55} & \ding{55} & --
& 0.682 & 42.0 \\

\textit{w/o WDL}
& \ding{51} & \ding{51} & \ding{55} & --
& \underline{0.692} & \underline{44.4} \\

\textit{w/o Weak Sel.}
& \ding{51} & \ding{51} & \ding{51} & \ding{55}
& 0.684 & 41.2 \\

\midrule

\textbf{StarTV-Agent}
& \ding{51} & \ding{51} & \ding{51} & \ding{51}
& \textbf{0.694} & \textbf{44.8} \\

\bottomrule
\end{tabular*}
\end{table}

\begin{table}[t]
\centering
\caption{Ablation of Lightweight Adaptive Workflow
Reuse (LAWR). ReAct and LAWR indicate available
execution paths; the full model selects one path
per task. Bold and underline indicate the best
and second-best results, respectively.}
\label{tab:inference_ablation}

\scriptsize
\setlength{\tabcolsep}{1pt}
\renewcommand{\arraystretch}{1.12}

\begin{tabular*}{\columnwidth}{
@{\extracolsep{\fill}}lcccc@{}}
\toprule

\textbf{Variant}
& \textbf{ReAct}
& \textbf{LAWR}
& \shortstack{\textbf{RoadSceneVQA}\\
\textbf{GPT-Score} $\uparrow$}
& \shortstack{\textbf{Traffic-VQA}\\
\textbf{OA} $\uparrow$} \\

\midrule

\textit{VLM-only}
& \ding{55} & \ding{55}
& \underline{0.643} & 42.3 \\

\textit{w/o LAWR}
& \ding{51} & \ding{55}
& \textbf{0.694} & \underline{44.8} \\

\midrule

\textbf{StarTV-Agent}
& \ding{51} & \ding{51}
& \textbf{0.694} & \textbf{45.7} \\

\bottomrule
\end{tabular*}
\end{table}

\subsection{Experimental Setups}
\label{sec:experimental_setups}

\noindent\textbf{Datasets.}
We evaluate StarTV-Agent on three traffic scenario understanding benchmarks: TUMTraf VideoQA~\cite{zhou2025tumtraf}, Traffic-VQA~\cite{zhang2026trafficvqa}, and RoadSceneVQA~\cite{guan2026roadscenevqa}. 
We follow the original dataset splits and evaluation
metrics for each benchmark.
Further benchmark and evaluation details are provided in Appendix~\ref{app:datasets}.

\noindent\textbf{Baselines.}
We compare StarTV-Agent against three groups of baselines.
\emph{Frontier-scale VLMs} include
DeepSeek-V4.1-Flash~\cite{deepseek2026v41flash},
Kimi-K3~\cite{moonshot2026kimik3},
Qwen3.8-Max~\cite{alibaba2026qwen38max},
and GPT-5.5~\cite{openai2026gpt55}.
\emph{Comparable-scale VLMs} comprise
Qwen3-VL-8B-Instruct~\cite{bai2025qwen3vl},
STEP3-VL-10B~\cite{huang2026step3vl},
and Qwen3.8-27B~\cite{qwen2026qwen38}.
\emph{Tool-augmented agents} include
DeepEyes~\cite{zheng2026deepeyes},
VTool-R1~\cite{wu2025vtoolr1},
and TAU-Agent~\cite{lin2026tauagent}.
This grouping enables comparisons with frontier models,
similarly sized VLMs, and existing tool-based approaches.
Implementation details and adaptations of the tool-augmented
baselines to our VQA tasks are provided in
Appendix~\ref{app:baselines}.

\noindent\textbf{Implementation Details.}
StarTV-Agent uses Qwen3-VL-8B-Instruct as the training backbone, Qwen3.8-27B for process scoring for RL, and Kimi-K3 for teacher demonstrations for SFT. Training and inference run on 8 $\times$ NVIDIA A800 GPUs, with Agent Lightning~\cite{luo2025agentlightning} coordinating rollouts, VERL~\cite{sheng2025hybridflow} optimizing the policy, and LangGraph~\cite{langgraph2026} orchestrating inference. RL updates all backbone parameters, while SFT uses LoRA~\cite{hu2022lora}. Each configuration is evaluated in a single run with a fixed seed of 42. We use the final trained checkpoint for the main comparisons in
Tables~\ref{tab:tumtraf_results}--\ref{tab:roadscene_results}.
Each benchmark starts with an empty, isolated experience store. 
For RSPO, we set $\alpha=0.8$ and $\beta=0.2$.
For LAWR, we set $a_{\min}=0.4$,
$s_{\min}=-0.10$, $K_S=3$,
$N_{\min}=2$, $\gamma_{\mathrm{prom}}=0.6$,
and $K_{\mathrm{rev}}=3$.
The reference cost uses an exponential moving
average with a smoothing factor of $0.3$.
 Further details are provided in Appendices~\ref{app:baselines} and~\ref{app:training_data}.

\subsection{Main Results}
\label{sec:main_results}

\noindent\textbf{TUMTraf VideoQA.}
We evaluate video understanding across ten question
categories.
As shown in Table~\ref{tab:tumtraf_results},
StarTV-Agent achieves the highest overall accuracy
among non-frontier models (58.3\%), outperforming
its Qwen3-VL-8B-Instruct backbone
and TAU-Agent by 2.4 and 3.3
percentage points, respectively.
Notably, it reaches 83.8\% on Easy counting,
surpassing even frontier-scale VLMs, and 71.7\%
on Hard existence, exceeding the strongest
non-frontier baseline by 9.1 points.
These results show that StarTV-Agent maintains
strong fine-grained traffic understanding while
supporting efficient workflow reuse.

\noindent\textbf{Traffic-VQA.}
We evaluate three visual settings: optical (OPT),
thermal infrared (TIR), and paired multimodal
(MUL).
As shown in Table~\ref{tab:traffic_results},
StarTV-Agent achieves the highest mean accuracy
among non-frontier models at 45.7\%, exceeding
its backbone and TAU-Agent by 3.4 and 2.6
percentage points, respectively.
It ranks first on OPT and TIR, with a notable
3.5-point gain over the strongest non-frontier
baseline on TIR.
On MUL, StarTV-Agent ranks second at 47.2\%,
improving upon its backbone by 3.5 points.
These results demonstrate strong performance
across different visual modalities.

\noindent\textbf{RoadSceneVQA.}
We further evaluate open-ended traffic question
answering using lexical metrics and GPT-based
scoring.
As shown in Table~\ref{tab:roadscene_results},
StarTV-Agent achieves a GPT-Score of 0.694,
ranking second among non-frontier models and
improving upon its backbone by 0.051.
This result approaches frontier-scale
DeepSeek-V4.1-Flash
(0.703) and Kimi-K3
(0.706), despite using an 8B backbone.
This competitive performance suggests that
efficient workflow reuse can coexist with
strong specialist-assisted reasoning for
open-ended traffic understanding.

\subsection{Ablation Studies}
\label{sec:ablations}

\noindent\textbf{Policy Learning.}
Table~\ref{tab:training_ablation} examines how
RSPO and WDL strengthen the slow reasoning path.
\textit{Qwen3-VL-8B-Instruct} is the untrained backbone
equipped with ReAct-based tool orchestration.
\textit{Terminal-only RL}
uses a GSPO-style objective~\cite{zheng2025gspo}
with terminal rewards alone.
StarTV-Agent~(\textit{w/o WDL}) adds RSPO but
excludes demonstration SFT, while
StarTV-Agent~(\textit{w/o Weak Sel.}) replaces
weakness-based question selection with balanced
random sampling for teacher demonstration generation.

On Traffic-VQA, RSPO improves accuracy from
42.0\% to 44.4\%, while adding WDL further
increases it to 44.8\%. Replacing weakness-based
selection with balanced random sampling reduces
accuracy to 41.2\% and RoadSceneVQA GPT-Score
from 0.694 to 0.684.
These results show that reliable slow-path
reasoning benefits from both role-sensitive
process feedback and demonstrations targeting
the student's weaknesses.

\noindent\textbf{Experience Evolution.}
Table~\ref{tab:inference_ablation} compares three
inference settings using the same Qwen3-VL-8B-Instruct backbone 
without additional training, isolating the effects of LAWR from policy learning.
\textit{VLM-only} answers questions without tools,
while StarTV-Agent~(\textit{w/o LAWR}) uses the slow
ReAct-based path~\cite{yao2023react} without
workflow reuse. StarTV-Agent denotes the full inference configuration that 
enables both the slow path and the fast path.

\begin{figure}[t]
    \centering
    \includegraphics[width=0.8\linewidth]{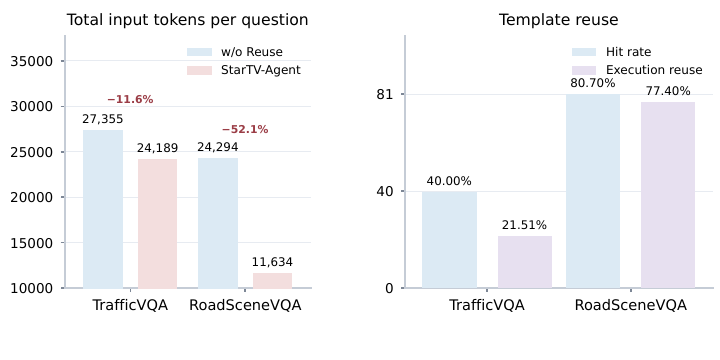}
    \caption{Input-token efficiency and workflow reuse on Traffic-VQA and RoadSceneVQA. Left: average input tokens per question with and without workflow reuse. Right: workflow retrieval hit and direct execution reuse rates.}
    \label{fig:deepdive}
\end{figure}

On Traffic-VQA, StarTV-Agent~(\textit{w/o LAWR}) improves accuracy
from 42.3\% to 44.8\%, while StarTV-Agent further increases
it to 45.7\%. As shown in
Fig.~\ref{fig:deepdive}, StarTV-Agent also reduces average
input-token consumption by 11.6\% and 52.1\%
on the two benchmarks, respectively.
These results demonstrate the independent benefits
of inference-time tool orchestration and workflow
reuse, without relying on additional policy training.

\subsection{Deep Dive}

\noindent\textbf{Token efficiency.}
Figure~\ref{fig:deepdive} compares average input
tokens per question with and without workflow reuse.
On Traffic-VQA, reuse reduces token consumption
from 27,355 to 24,189 (11.6\%) while improving
accuracy by 0.9 percentage points
(Table~\ref{tab:inference_ablation}).
On RoadSceneVQA, input tokens decrease from
24,294 to 11,634 (52.1\%) while GPT-Score
remains unchanged at 0.694.
These findings show that the fast workflow path
can avoid repeated planning while preserving
the benefits of deliberate tool reasoning.

\noindent\textbf{Workflow reuse behavior.}
Figure~\ref{fig:deepdive} also reports workflow
retrieval hit rates and actual workflow reuse rates.
On Traffic-VQA, 40.00\% of questions retrieve
a matching workflow, while 21.51\% directly reuse one.
On RoadSceneVQA, these rates reach 80.70\% and
77.40\%, respectively.
The gap between retrieval and execution shows
that the agent does not blindly follow retrieved
workflows: only suitable workflows enter the fast
path, while other tasks use slow ReAct-based
orchestration.
The higher reuse rate on RoadSceneVQA also
coincides with greater token savings, suggesting
that the benefit of the fast path grows when
more tasks share reusable tool-use patterns.

\begin{figure}
    \centering
    \includegraphics[width=0.8\linewidth]{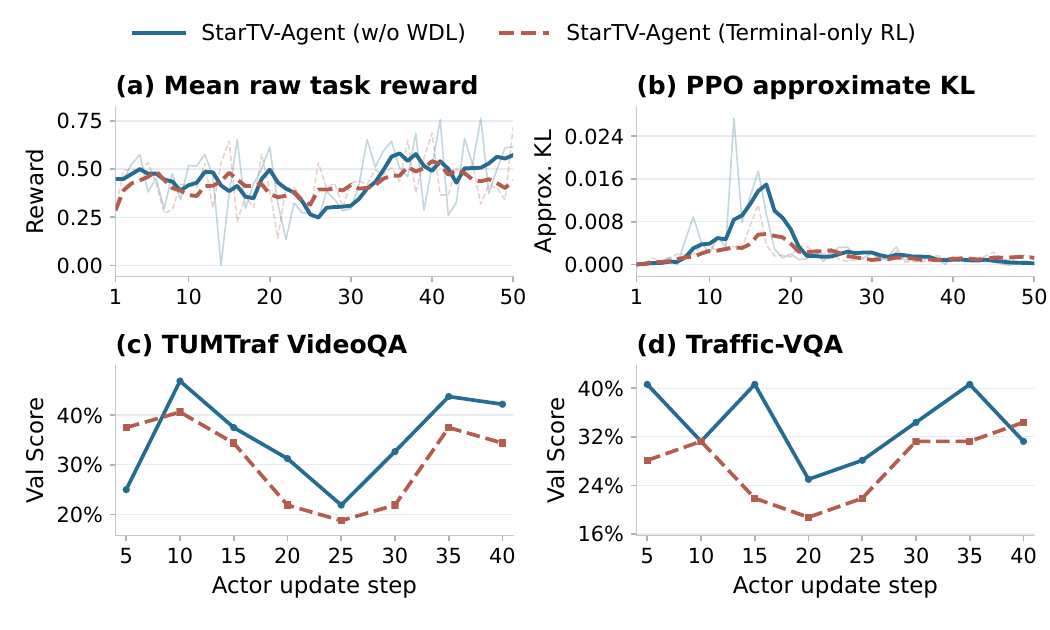}
    
    \caption{RL training diagnostics comparing
    StarTV-Agent (w/o WDL) with terminal-only RL.
    (a) Mean raw task reward during training.
    (b) PPO approximate KL divergence.
    (c--d) Validation performance on TUMTraf VideoQA
    and Traffic-VQA.
    Solid and dashed lines represent the two methods,
    while faint lines in (a--b) show raw measurements
    and bold lines show smoothed trends.}
    \label{fig:rl_diagnostics}

\end{figure}

\noindent\textbf{RL training diagnostics.}
Figure~\ref{fig:rl_diagnostics} compares StarTV-Agent~(\textit{w/o WDL}) with StarTV-Agent~(\textit{Terminal-only RL})
throughout training. StarTV-Agent~(\textit{w/o WDL}) shows an overall upward
trend in mean task reward, despite fluctuations
in both methods. Its approximate KL rises
temporarily during early updates but gradually
returns to near zero, indicating that the larger
initial policy changes do not persist.
On both TUMTraf VideoQA and Traffic-VQA, StarTV-Agent~(\textit{w/o WDL})
achieves higher validation scores at most
evaluated checkpoints.
These results suggest that role-sensitive
process advantages provide more useful
learning signals than terminal rewards alone,
leading to better generalization across
traffic understanding tasks.

\noindent\textbf{Stage-wise cost analysis.}
Table~\ref{tab:cost_breakdown} shows why avoiding
repeated slow-path planning matters.
On Traffic-VQA, the four decision stages from
Planner to Verifier consume 23,463 input tokens
and 12.269 seconds per question.
On RoadSceneVQA, these stages consume 11,110
input tokens and 4.408 seconds, with the Verifier
alone accounting for 6,945 input tokens.
In contrast, the workflow reuse step itself
takes only 0.047 and 0.014 seconds on the two
benchmarks, respectively.
By following stored tool sequences, the fast
path avoids repeated planning, tool selection,
and intermediate verification.
Experience maintenance takes 3.649 and
2.439 seconds, but runs only after the final
answer is submitted.
Together, these results explain how StarTV-Agent
balances the flexibility of slow reasoning with
the efficiency of fast workflow execution.

\begin{table}[t]
\centering
\caption{Stage-wise inference cost on Traffic-VQA
and RoadSceneVQA. We report average input tokens,
output tokens, and execution time per question
for each component. ``--'' indicates stages
without VLM token consumption.}
\label{tab:cost_breakdown}

\scriptsize
\setlength{\tabcolsep}{3pt}
\renewcommand{\arraystretch}{1.08}

\begin{tabular*}{\columnwidth}{
@{\extracolsep{\fill}}lrrr@{}}
\toprule
Stage & Avg. Input Tokens & Avg. Output Tokens & Time (s) \\
\midrule
\multicolumn{4}{l}{\textit{Traffic-VQA}} \\
\midrule
Planner                & 4,972 & 166 & 2.393 \\
Tool Selector          & 7,008 & 337 & 4.598 \\
Executor               & 4,866 & 53  & 0.984 \\
Verifier               & 6,617 & 302 & 4.294 \\
Summary                & 573   & 55  & 0.720 \\
Tool Execution         & --    & --  & 6.801 \\
Workflow Reuse         & --    & --  & 0.047 \\
Experience Maintenance & 2,401 & 284 & 3.649 \\
\midrule
\multicolumn{4}{l}{\textit{RoadSceneVQA}} \\
\midrule
Planner                & 1,459 & 41  & 0.674 \\
Tool Selector          & 1,548 & 41  & 0.646 \\
Executor               & 1,158 & 8   & 0.191 \\
Verifier               & 6,945 & 158 & 2.897 \\
Summary                & 505   & 96  & 1.193 \\
Tool Execution         & --    & --  & 4.026 \\
Workflow Reuse         & --    & --  & 0.014 \\
Experience Maintenance & 1,849 & 183 & 2.439 \\
\bottomrule
\end{tabular*}
\end{table}

\noindent\textbf{Real-World Generalization.}
To further assess practical generalization,
we evaluate StarTV-Agent on a private dataset
collected from a real-world traffic-police
deployment, comprising 1,999 binary VQA samples
across six traffic-violation categories.
StarTV-Agent achieves 75.9\% overall accuracy
and 79.0\% F1, outperforming the strongest
VLM baseline by 15.7\% and 6.1\%, respectively, in relative terms.
It also achieves the best performance on
four of the six categories, with notable gains
in seatbelt and red-light violation detection.
These findings further support the effectiveness
of StarTV-Agent in practical traffic-review
scenarios. Detailed results are presented in
Appendix~\ref{sec:private-client}.

\section{Conclusion}
\label{sec:conclusion}

We presented \textbf{StarTV-Agent}, a
generalist--specialist framework that integrates
slow reasoning with fast workflow reuse for
traffic scenario understanding.
RSPO and WDL improve multi-step tool orchestration
through role-sensitive process feedback and
targeted teacher demonstrations, while LAWR
enables efficient workflow reuse and label-free
experience evolution.
Experiments on three benchmarks demonstrate
improvements in task performance and inference
efficiency, achieving up to 8.0\% relative
performance gains and 52.1\% input-token savings.
StarTV-Agent further connects adaptive reasoning
with reusable experience.
The slow path supports dynamic planning and
error recovery, while the fast path avoids
redundant decisions by reusing established
tool-use patterns.
Nevertheless, experience complements rather
than replaces reasoning: workflows must remain
applicable to new inputs, and execution feedback
does not necessarily reflect answer correctness.
Future work will explore richer feedback signals
and more flexible workflow adaptation across tasks.

\section*{LLM Usage Statement}

We used ChatGPT (OpenAI) to assist with language
editing and the organization of the manuscript,
including the abstract, introduction, related work,
methodology, and figure descriptions.
All technical claims, experimental results, and
final manuscript content were reviewed and
verified by the authors.

\clearpage
\bibliographystyle{plainnat}
\bibliography{main}

\begin{thebibliography}{63}
\providecommand{\natexlab}[1]{#1}
\providecommand{\url}[1]{\texttt{#1}}
\expandafter\ifx\csname urlstyle\endcsname\relax
  \providecommand{\doi}[1]{doi: #1}\else
  \providecommand{\doi}{doi: \begingroup \urlstyle{rm}\Url}\fi

\bibitem[{Alibaba Cloud}(2026)]{alibaba2026qwen38max}
{Alibaba Cloud}.
\newblock {Qwen3.8-Max Model Documentation}.
\newblock \url{https://help.aliyun.com/en/model-studio/qwen3-8-max}, 2026.
\newblock Accessed 2026-11-09.

\bibitem[Bai et~al.(2025)]{bai2025qwen3vl}
Shuai Bai et~al.
\newblock {Qwen3-VL Technical Report}.
\newblock \emph{arXiv preprint arXiv:2511.21631}, 2025.
\newblock URL \url{https://arxiv.org/abs/2511.21631}.

\bibitem[Cai et~al.(2026)Cai, Guo, Huang, Du, Huang, Huang, Ma, Hu, Zeng, Tang,
  and Sun]{cai2026memopilot}
Yishuo Cai, Xingyu Guo, Xuancheng Huang, Jinhua Du, Can Huang, Wenxuan Huang,
  Wenhan Ma, Yuyang Hu, Aohan Zeng, Jie Tang, and Xu~Sun.
\newblock {From Player to Master: Enhancing Test-Time Learning of {LLM} Agents
  via Reinforcement Learning over Memory}.
\newblock In \emph{Proceedings of the 43rd International Conference on Machine
  Learning}, volume 306, pages 10503--10520. PMLR, 2026.
\newblock URL \url{https://proceedings.mlr.press/v306/cai26b.html}.

\bibitem[Chen et~al.(2026)Chen, Wang, Yue, Yan, Yu, Huang, Liu, Wen, Chen, Liu,
  Li, and Wang]{chen2026videochatm1}
Boyu Chen, Zikang Wang, Zhengrong Yue, Kainan Yan, Chenyun Yu, Yi~Huang, Zijun
  Liu, Yafei Wen, Xiaoxin Chen, Yang Liu, Peng Li, and Yali Wang.
\newblock {VideoChat-M1: Collaborative Policy Planning for Video Understanding
  via Multi-Agent Reinforcement Learning}.
\newblock In \emph{Proceedings of the IEEE/CVF Conference on Computer Vision
  and Pattern Recognition (CVPR)}, pages 33772--33783, 2026.
\newblock URL
  \url{https://openaccess.thecvf.com/content/CVPR2026/html/Chen_VideoChat-M1_Collaborative_Policy_Planning_for_Video_Understanding_via_Multi-Agent_Reinforcement_CVPR_2026_paper.html}.

\bibitem[Chen et~al.(2025)Chen, Li, Zhang, Liu, Li, Gao, Hong, Tian, Zhao, Li,
  Yeung, Lu, and Jia]{chen2025codalm}
Kai Chen, Yanze Li, Wenhua Zhang, Yanxin Liu, Pengxiang Li, Ruiyuan Gao,
  Lanqing Hong, Meng Tian, Xinhai Zhao, Zhenguo Li, Dit-Yan Yeung, Huchuan Lu,
  and Xu~Jia.
\newblock {Automated Evaluation of Large Vision-Language Models on Self-Driving
  Corner Cases}.
\newblock In \emph{Proceedings of the Winter Conference on Applications of
  Computer Vision (WACV)}, pages 7806--7815, 2025.
\newblock URL
  \url{https://openaccess.thecvf.com/content/WACV2025/html/Chen_Automated_Evaluation_of_Large_Vision-Language_Models_on_Self-Driving_Corner_Cases_WACV_2025_paper.html}.

\bibitem[Cheng et~al.(2026)Cheng, He, Feng, Xu, Yan, Feng, and
  An]{cheng2026graphgpo}
Xin Cheng, Shuo He, Lang Feng, Haiyang Xu, Ming Yan, Lei Feng, and Bo~An.
\newblock {Beyond Trajectory-Level Attribution: Graph-Based Credit Assignment
  for Agentic Reinforcement Learning}.
\newblock In \emph{Proceedings of the 43rd International Conference on Machine
  Learning}, volume 306, pages 18861--18886. PMLR, 2026.
\newblock URL \url{https://proceedings.mlr.press/v306/cheng26e.html}.

\bibitem[{DeepSeek}(2026)]{deepseek2026v41flash}
{DeepSeek}.
\newblock {DeepSeek-V4.1-Flash}.
\newblock \url{https://api-docs.deepseek.com/news/news260910/}, 2026.
\newblock Official release documentation; accessed 2026-11-09.

\bibitem[Ding et~al.(2026)Ding, Fang, Liu, Zang, Cao, Zhao, Duan, Dong, Liang,
  Wang, He, Lin, and Wang]{ding2026armthinker}
Shengyuan Ding, Xinyu Fang, Ziyu Liu, Yuhang Zang, Yuhang Cao, Xiangyu Zhao,
  Haodong Duan, Xiaoyi Dong, Jianze Liang, Bin Wang, Conghui He, Dahua Lin, and
  Jiaqi Wang.
\newblock {ARM-Thinker: Reinforcing Multimodal Generative Reward Models with
  Agentic Tool Use and Visual Reasoning}.
\newblock In \emph{Proceedings of the IEEE/CVF Conference on Computer Vision
  and Pattern Recognition (CVPR)}, pages 22195--22205, 2026.
\newblock URL
  \url{https://openaccess.thecvf.com/content/CVPR2026/html/Ding_ARM-Thinker_Reinforcing_Multimodal_Generative_Reward_Models_with_Agentic_Tool_Use_CVPR_2026_paper.html}.

\bibitem[Feng et~al.(2025)Feng, Xue, Liu, and An]{feng2025gigpo}
Lang Feng, Zhenghai Xue, Tingcong Liu, and Bo~An.
\newblock {Group-in-Group Policy Optimization for LLM Agent Training}.
\newblock In \emph{Advances in Neural Information Processing Systems},
  volume~38, pages 46375--46408, 2025.
\newblock \doi{10.52202/085713-1544}.
\newblock URL
  \url{https://proceedings.neurips.cc/paper_files/paper/2025/hash/420c9f777c0b4f78d515e53cf74d58b2-Abstract-Conference.html}.

\bibitem[Feng et~al.(2026)Feng, Luo, Xu, Hua, Xie, Yang, Liu, and
  You]{feng2026expweaver}
Tao Feng, Tianyang Luo, Jingjun Xu, Zhigang Hua, Yan Xie, Shuang Yang, Ge~Liu,
  and Jiaxuan You.
\newblock {{E}xp{W}eaver: {LLM} Agents Learn from Experience via Latent {RAG}}.
\newblock In \emph{Proceedings of the 43rd International Conference on Machine
  Learning}, volume 306, pages 30156--30220. PMLR, 2026.
\newblock URL \url{https://proceedings.mlr.press/v306/feng26i.html}.

\bibitem[Gao et~al.(2026{\natexlab{a}})Gao, Li, Jia, Yuan, Sun, Sun, and
  Li]{gao2026sapo}
Lei Gao, Zhuoming Li, Mengxi Jia, Jiakang Yuan, Hongbo Sun, Hao Sun, and
  Xuelong Li.
\newblock Segment-aligned policy optimization for multi-modal reasoning.
\newblock In \emph{Proceedings of the 43rd International Conference on Machine
  Learning}, volume 306 of \emph{Proceedings of Machine Learning Research},
  pages 33285--33299. PMLR, 2026{\natexlab{a}}.
\newblock URL \url{https://proceedings.mlr.press/v306/gao26n.html}.

\bibitem[Gao et~al.(2026{\natexlab{b}})Gao, Fang, Ding, Li, Wei, Sun, Dong,
  Yuan, Xu, Liang, Sun, and Xin]{gao2026motto}
Tianyi Gao, Han Fang, Tianyi Ding, Hao Li, Xin Wei, Hongbo Sun, Xiaodong Dong,
  Ye~Yuan, Jinglin Xu, Kongming Liang, Hao Sun, and Jingmin Xin.
\newblock {Mixture-of-Thought-Tokens}: Unifying perception and reasoning for
  free-form multimodal grounding.
\newblock In \emph{Proceedings of the 34th ACM International Conference on
  Multimedia}, pages 2295--2304. Association for Computing Machinery,
  2026{\natexlab{b}}.
\newblock \doi{10.1145/3767308.3835134}.

\bibitem[Guan et~al.(2026)Guan, Hu, Chen, Xiao, Xia, Liu, Chen, Tang, Ouyang,
  Liang, Fan, Sun, and Yue]{guan2026roadscenevqa}
Runwei Guan, Rongsheng Hu, Shangshu Chen, Ningyuan Xiao, Xue Xia, Jiayang Liu,
  Beibei Chen, Ziren Tang, Ningwei Ouyang, Shaofeng Liang, Yuxuan Fan, Wanjie
  Sun, and Yutao Yue.
\newblock {RoadSceneVQA: Benchmarking Visual Question Answering in Roadside
  Perception Systems for Intelligent Transportation System}.
\newblock In \emph{Proceedings of the AAAI Conference on Artificial
  Intelligence}, volume~40, pages 4366--4375. {AAAI} Press, 2026.
\newblock \doi{10.1609/AAAI.V40I6.42434}.
\newblock URL \url{https://doi.org/10.1609/aaai.v40i6.42434}.

\bibitem[Guo et~al.(2026)Guo, Hong, Zhang, Jia, and Jin]{guo2026codevision}
Zirun Guo, Minjie Hong, Feng Zhang, Kai Jia, and Tao Jin.
\newblock {Thinking with Programming Vision: Towards a Unified View for
  Thinking with Images}.
\newblock In \emph{Proceedings of the IEEE/CVF Conference on Computer Vision
  and Pattern Recognition (CVPR)}, pages 33467--33476, 2026.
\newblock URL
  \url{https://openaccess.thecvf.com/content/CVPR2026/html/Guo_Thinking_with_Programming_Vision_Towards_a_Unified_View_for_Thinking_CVPR_2026_paper.html}.

\bibitem[Hu et~al.(2022)Hu, Shen, Wallis, Allen-Zhu, Li, Wang, Wang, and
  Chen]{hu2022lora}
Edward~J. Hu, Yelong Shen, Phillip Wallis, Zeyuan Allen-Zhu, Yuanzhi Li, Shean
  Wang, Lu~Wang, and Weizhu Chen.
\newblock {LoRA}: Low-rank adaptation of large language models.
\newblock In \emph{International Conference on Learning Representations}, 2022.
\newblock URL \url{https://openreview.net/forum?id=nZeVKeeFYf9}.

\bibitem[Huang et~al.(2026{\natexlab{a}})]{huang2026step3vl}
Ailin Huang et~al.
\newblock {STEP3-VL-10B Technical Report}.
\newblock \emph{arXiv preprint arXiv:2601.09668}, 2026{\natexlab{a}}.
\newblock URL \url{https://arxiv.org/abs/2601.09668}.

\bibitem[Huang et~al.(2026{\natexlab{b}})Huang, Chen, Xu, Li, Xu, Chen, Liu,
  Kong, and Zhu]{huang2026meta}
Jing Huang, Luyuan Chen, Zhijie Xu, Yadong Li, Xingzhong Xu, Siye Chen, Jie
  Liu, Ming Kong, and Qiang Zhu.
\newblock {META: Meta Evolution of Tool Trajectory Adaptation for Long-Video
  Understanding}.
\newblock In \emph{Proceedings of the IEEE/CVF Conference on Computer Vision
  and Pattern Recognition (CVPR)}, pages 9837--9846, 2026{\natexlab{b}}.
\newblock URL
  \url{https://openaccess.thecvf.com/content/CVPR2026/html/Huang_META_Meta_Evolution_of_Tool_Trajectory_Adaptation_for_Long-Video_Understanding_CVPR_2026_paper.html}.

\bibitem[Huang et~al.(2026{\natexlab{c}})Huang, Jia, Sun, Li, and
  Han]{huang2026calibrl}
Zhuoxu Huang, Mengxi Jia, Hao Sun, Xuelong Li, and Jungong Han.
\newblock Controllable exploration in hybrid-policy {RLVR} for multi-modal
  reasoning.
\newblock In \emph{The Fourteenth International Conference on Learning
  Representations}, 2026{\natexlab{c}}.
\newblock URL \url{https://openreview.net/forum?id=5wxyCidRsK}.

\bibitem[Janjua et~al.(2026)Janjua, Silva, Niu, and
  Rashidi]{janjua2026perceptionprograms}
Muhammad~Kamran Janjua, Hugo Silva, Di~Niu, and Bahador Rashidi.
\newblock {Don't Show Pixels, Show Cues: Unlocking Visual Tool Reasoning in
  Language Models via Perception Programs}.
\newblock In \emph{Proceedings of the IEEE/CVF Conference on Computer Vision
  and Pattern Recognition (CVPR)}, pages 5165--5174, 2026.
\newblock URL
  \url{https://openaccess.thecvf.com/content/CVPR2026/html/Janjua_Dont_Show_Pixels_Show_Cues_Unlocking_Visual_Tool_Reasoning_in_CVPR_2026_paper.html}.

\bibitem[Jiang et~al.(2026)Jiang, Su, Qu, and Fung]{jiang2026xskill}
Guanyu Jiang, Zhaochen Su, Xiaoye Qu, and Yi~R. Fung.
\newblock {{XS}kill: Continual Learning from Experience and Skills in
  Multimodal Agents}.
\newblock In \emph{Proceedings of the 43rd International Conference on Machine
  Learning}, volume 306, pages 53078--53104. PMLR, 2026.
\newblock URL \url{https://proceedings.mlr.press/v306/jiang26aq.html}.

\bibitem[Ke et~al.(2025)Ke, G, Leng, Cai, Khan, Wang, Haghighi, Rezatofighi,
  and Chandraker]{ke2025dwim}
Fucai Ke, Vijay Kumar~B G, Xingjian Leng, Zhixi Cai, Zaid Khan, Weiqing Wang,
  Pari~Delir Haghighi, Hamid Rezatofighi, and Manmohan Chandraker.
\newblock {DWIM: Towards Tool-aware Visual Reasoning via Discrepancy-aware
  Workflow Generation \& Instruct-Masking Tuning}.
\newblock In \emph{Proceedings of the IEEE/CVF International Conference on
  Computer Vision (ICCV)}, pages 3378--3389, 2025.
\newblock URL
  \url{https://openaccess.thecvf.com/content/ICCV2025/html/Ke_DWIM_Towards_Tool-aware_Visual_Reasoning_via_Discrepancy-aware_Workflow_Generation__ICCV_2025_paper.html}.

\bibitem[{LangChain}(2026)]{langgraph2026}
{LangChain}.
\newblock {LangGraph: A Framework for Building Stateful Agents}.
\newblock \url{https://github.com/langchain-ai/langgraph}, 2026.
\newblock Software repository; accessed 2026-11-09.

\bibitem[Li et~al.(2026{\natexlab{a}})Li, Zheng, Song, Liu, Yang, Sun, Zhong,
  Zhang, Guo, Fang, and Wang]{li2026guidedopd}
Gengsheng Li, Mao Zheng, Mingyang Song, Ruiqi Liu, Tianyu Yang, Jie Sun, Qiyong
  Zhong, Dan Zhang, Haiyun Guo, Junfeng Fang, and Jinqiao Wang.
\newblock {On-Policy Distillation with Curriculum Turn-level Guidance for
  Multi-turn Agents}.
\newblock In \emph{Proceedings of the 2026 Conference on Empirical Methods in
  Natural Language Processing}, pages 32539--32555. Association for
  Computational Linguistics, 2026{\natexlab{a}}.
\newblock \doi{10.18653/2026.emnlp-main.1424}.
\newblock URL \url{https://aclanthology.org/2026.emnlp-main.1424/}.

\bibitem[Li et~al.(2026{\natexlab{b}})Li, Fang, Pan, Wei, Sun, Xu, Lin, Yuan,
  He, Yu, and Sun]{li2026geoanchor}
Hao Li, Han Fang, Zixin Pan, Xin Wei, Hongbo Sun, Jinglin Xu, Zhiyu Lin,
  Ye~Yuan, Zhongjiang He, Yu~Yu, and Hao Sun.
\newblock {GeoAnchor}: Collaborative reasoning via latent decomposition for
  {3D} spatial understanding.
\newblock In \emph{Proceedings of the 34th ACM International Conference on
  Multimedia}, pages 4603--4612. Association for Computing Machinery,
  2026{\natexlab{b}}.
\newblock \doi{10.1145/3767308.3835036}.

\bibitem[Li et~al.(2026{\natexlab{c}})Li, Li, Shen, Liu, Chang, and
  Shan]{li2026lenswalk}
Keliang Li, Yansong Li, Hongze Shen, Mengdi Liu, Hong Chang, and Shiguang Shan.
\newblock {LensWalk: Agentic Video Understanding by Planning How You See in
  Videos}.
\newblock In \emph{Proceedings of the IEEE/CVF Conference on Computer Vision
  and Pattern Recognition (CVPR)}, pages 19518--19528, 2026{\natexlab{c}}.
\newblock URL
  \url{https://openaccess.thecvf.com/content/CVPR2026/html/Li_LensWalk_Agentic_Video_Understanding_by_Planning_How_You_See_in_CVPR_2026_paper.html}.

\bibitem[Li et~al.(2026{\natexlab{d}})Li, Zhang, Han, Liu, Xie, Zhang, Choi,
  Zou, and Lu]{li2026agentflow}
Zhuofeng Li, Haoxiang Zhang, Seungju Han, Sheng Liu, Jianwen Xie, Yu~Zhang,
  Yejin Choi, James Zou, and Pan Lu.
\newblock {In-the-Flow Agentic System Optimization for Effective Planning and
  Tool Use}.
\newblock In \emph{International Conference on Learning Representations
  (ICLR)}, 2026{\natexlab{d}}.
\newblock URL \url{https://agentflow.stanford.edu/}.

\bibitem[Lin et~al.(2026)Lin, Shi, Lockyer, Madabushi, Evans, Li, Wang, and
  Zhang]{lin2026tauagent}
Yuqiang Lin, Yan Shi, Sam Lockyer, Harish~Tayyar Madabushi, Adrian Evans,
  Wenbin Li, Yinhai Wang, and Nic Zhang.
\newblock {TAU-Agent: An Agentic Retrieval-Augmented Framework for Traffic
  Anomaly Understanding}.
\newblock \emph{arXiv preprint arXiv:2608.25935}, 2026.
\newblock URL \url{https://arxiv.org/abs/2608.25935}.

\bibitem[Liu et~al.(2026)Liu, Chai, Zhu, Tang, Ye, Ma, Zhang, Bai, and
  Chen]{liu2026mlagent}
Zexi Liu, Jingyi Chai, Xinyu Zhu, Shuo Tang, Rui Ye, Weiyu Ma, Bo~Zhang, Lei
  Bai, and Siheng Chen.
\newblock {{ML}-Agent: Reinforcing {LLM} Agents for Autonomous Machine Learning
  Engineering}.
\newblock In \emph{Proceedings of the 43rd International Conference on Machine
  Learning}, volume 306, pages 75549--75579. PMLR, 2026.
\newblock URL \url{https://proceedings.mlr.press/v306/liu26i.html}.

\bibitem[Lu et~al.(2026{\natexlab{a}})Lu, Xu, Fang, Zhang, Yu, Srivastava,
  Zhuang, Elhoseiny, Fleming, Yang, Tu, Xie, Xiao, Jin, Shi, and
  Wang]{lu2026vistagym}
Meng Lu, Ran Xu, Yi~Fang, Wenxuan Zhang, Yue Yu, Gaurav Srivastava, Yuchen
  Zhuang, Mohamed Elhoseiny, Charles Fleming, Carl Yang, Zhengzhong Tu, Yang
  Xie, Guanghua Xiao, Di~Jin, Wenqi Shi, and Xuan Wang.
\newblock {Scaling Agentic Reinforcement Learning for Tool-Integrated Reasoning
  in VLMs}.
\newblock In \emph{Proceedings of the IEEE/CVF Conference on Computer Vision
  and Pattern Recognition (CVPR)}, pages 26518--26529, 2026{\natexlab{a}}.
\newblock URL
  \url{https://openaccess.thecvf.com/content/CVPR2026/html/Lu_Scaling_Agentic_Reinforcement_Learning_for_Tool-Integrated_Reasoning_in_VLMs_CVPR_2026_paper.html}.

\bibitem[Lu et~al.(2026{\natexlab{b}})Lu, Chen, Liu, Thapa, Boen, and
  Zou]{lu2026octotools}
Pan Lu, Bowen Chen, Sheng Liu, Rahul Thapa, Joseph Boen, and James Zou.
\newblock {OctoTools}: A multi-agent framework with extensible tools for
  complex reasoning.
\newblock In \emph{Proceedings of the 64th Annual Meeting of the Association
  for Computational Linguistics (Volume 1: Long Papers)}, pages 1--86.
  Association for Computational Linguistics, 2026{\natexlab{b}}.
\newblock \doi{10.18653/v1/2026.acl-long.1}.
\newblock URL \url{https://aclanthology.org/2026.acl-long.1/}.

\bibitem[Luo et~al.(2025)Luo, Zhang, He, Wang, Zhao, Li, Qiu, and
  Yang]{luo2025agentlightning}
Xufang Luo, Yuge Zhang, Zhiyuan He, Zilong Wang, Siyun Zhao, Dongsheng Li,
  Luna~K. Qiu, and Yuqing Yang.
\newblock {Agent Lightning: Train ANY AI Agents with Reinforcement Learning}.
\newblock \emph{arXiv preprint arXiv:2508.03680}, 2025.
\newblock URL \url{https://arxiv.org/abs/2508.03680}.

\bibitem[Ma et~al.(2026)Ma, Zhang, Zheng, Tang, Li, Lu, Zhou, Zhang, Zhan,
  Zhan, Hao, Lang, and Yu]{ma2026drivecombo}
Enhui Ma, Jiahuan Zhang, Guantian Zheng, Tao Tang, Shengbo~Eben Li, Yuhang Lu,
  Xia Zhou, Xueyang Zhang, Yifei Zhan, Kun Zhan, Zhihui Hao, Xianpeng Lang, and
  Kaicheng Yu.
\newblock {DriveCombo: Benchmarking Compositional Traffic Rule Reasoning in
  Autonomous Driving}.
\newblock In \emph{Proceedings of the IEEE/CVF Conference on Computer Vision
  and Pattern Recognition (CVPR)}, pages 32113--32123, 2026.
\newblock URL
  \url{https://openaccess.thecvf.com/content/CVPR2026/html/Ma_DriveCombo_Benchmarking_Compositional_Traffic_Rule_Reasoning_in_Autonomous_Driving_CVPR_2026_paper.html}.

\bibitem[Mi et~al.(2026)Mi, Ma, Yang, Li, Wang, Zhang, and
  Wang]{mi2026skillpro}
Qirui Mi, Zhijian Ma, Mengyue Yang, Haoxuan Li, Yisen Wang, Haifeng Zhang, and
  Jun Wang.
\newblock {Skill-Pro: Learning Reusable Skills from Experience via
  Non-Parametric {PPO} for {LLM} Agents}.
\newblock In \emph{Proceedings of the 43rd International Conference on Machine
  Learning}, volume 306, pages 88574--88595. PMLR, 2026.
\newblock URL \url{https://proceedings.mlr.press/v306/mi26d.html}.

\bibitem[{Moonshot AI}(2026)]{moonshot2026kimik3}
{Moonshot AI}.
\newblock {Kimi K3 Model Card}.
\newblock \url{https://huggingface.co/moonshotai/Kimi-K3}, 2026.
\newblock Accessed 2026-11-09.

\bibitem[{OpenAI}(2026)]{openai2026gpt55}
{OpenAI}.
\newblock {GPT-5.5 Model Documentation}.
\newblock \url{https://developers.openai.com/api/docs/models/gpt-5.5}, 2026.
\newblock Accessed 2026-11-09.

\bibitem[Parikh et~al.(2025)Parikh, Rawat, T., Ghosh, and
  Sarvadevabhatla]{parikh2025roadsocial}
Chirag Parikh, Deepti Rawat, Rakshitha~R. T., Tathagata Ghosh, and Ravi~Kiran
  Sarvadevabhatla.
\newblock {RoadSocial: A Diverse VideoQA Dataset and Benchmark for Road Event
  Understanding from Social Video Narratives}.
\newblock In \emph{Proceedings of the IEEE/CVF Conference on Computer Vision
  and Pattern Recognition (CVPR)}, pages 19002--19011, 2025.
\newblock URL
  \url{https://openaccess.thecvf.com/content/CVPR2025/html/Parikh_RoadSocial_A_Diverse_VideoQA_Dataset_and_Benchmark_for_Road_Event_CVPR_2025_paper.html}.

\bibitem[Qiu et~al.(2026)Qiu, Xie, Huo, Tian, and Ye]{qiu2026longvideor1}
Jihao Qiu, Lingxi Xie, Xinyue Huo, Qi~Tian, and Qixiang Ye.
\newblock {LongVideo-R1: Smart Navigation for Low-cost Long Video
  Understanding}.
\newblock In \emph{Proceedings of the IEEE/CVF Conference on Computer Vision
  and Pattern Recognition (CVPR)}, pages 40505--40515, 2026.
\newblock URL
  \url{https://openaccess.thecvf.com/content/CVPR2026/html/Qiu_LongVideo-R1_Smart_Navigation_for_Low-cost_Long_Video_Understanding_CVPR_2026_paper.html}.

\bibitem[{Qwen Team}(2026{\natexlab{a}})]{qwen2026qwen38}
{Qwen Team}.
\newblock {Qwen3.8-27B Model Card}.
\newblock \url{https://huggingface.co/Qwen/Qwen3.8-27B}, 2026{\natexlab{a}}.
\newblock Accessed 2026-11-09.

\bibitem[{Qwen Team}(2026{\natexlab{b}})]{qwen2026qwen38flashnext}
{Qwen Team}.
\newblock {Qwen3.8-Flash-Next}: A new architecture, towards ultimate
  cost-efficiency.
\newblock Qwen Official Blog, August 2026{\natexlab{b}}.
\newblock URL \url{https://qwen.ai/blog?id=qwen3.8-flash-next}.

\bibitem[Ren et~al.(2026)Ren, Yue, Li, Wang, Xu, Liu, Pan, and
  Di]{ren2026flowevo}
Zeyu Ren, Ling Yue, Ran Li, Yishu Wang, Shengxiang Xu, Hanmo Liu, Shaowu Pan,
  and Shimin Di.
\newblock {FlowEvo}: Self-evolving agents through the co-evolution of workflows
  and executable skills.
\newblock In \emph{Third Conference on Language Modeling (COLM)}, 2026.
\newblock URL \url{https://openreview.net/forum?id=hU2N7IIkcE}.

\bibitem[Shao et~al.(2026)Shao, Yin, Lyu, Yu, Guo, Tsang, Kwok, and
  Li]{shao2026nsi}
Jie-Jing Shao, Haiyan Yin, Yueming Lyu, Xingrui Yu, Lan-Zhe Guo, Ivor Tsang,
  James Kwok, and Yu-Feng Li.
\newblock {Lifting Traces to Logic: Programmatic Skill Induction with
  Neuro-Symbolic Learning for Long-Horizon Agentic Tasks}.
\newblock In \emph{Proceedings of the 43rd International Conference on Machine
  Learning}, volume 306, pages 110314--110337. PMLR, 2026.
\newblock URL \url{https://proceedings.mlr.press/v306/shao26i.html}.

\bibitem[Sheng et~al.(2025)Sheng, Zhang, Ye, Wu, Zhang, Zhang, Peng, Lin, and
  Wu]{sheng2025hybridflow}
Guangming Sheng, Chi Zhang, Zilingfeng Ye, Xibin Wu, Wang Zhang, Ru~Zhang,
  Yanghua Peng, Haibin Lin, and Chuan Wu.
\newblock {HybridFlow: A Flexible and Efficient RLHF Framework}.
\newblock In \emph{Proceedings of the Twentieth European Conference on Computer
  Systems (EuroSys)}, pages 1279--1297. ACM, 2025.
\newblock \doi{10.1145/3689031.3696075}.

\bibitem[Shinn et~al.(2023)Shinn, Cassano, Gopinath, Narasimhan, and
  Yao]{shinn2023reflexion}
Noah Shinn, Federico Cassano, Ashwin Gopinath, Karthik Narasimhan, and Shunyu
  Yao.
\newblock {Reflexion: language agents with verbal reinforcement learning}.
\newblock In \emph{Advances in Neural Information Processing Systems},
  volume~36, pages 8634--8652, 2023.
\newblock \doi{10.52202/075280-0377}.
\newblock URL
  \url{https://papers.neurips.cc/paper_files/paper/2023/hash/1b44b878bb782e6954cd888628510e90-Abstract-Conference.html}.

\bibitem[Sima et~al.(2025)Sima, Renz, Chitta, Chen, Zhang, Xie, Bei{\ss}wenger,
  Luo, Geiger, and Li]{sima2024drivelm}
Chonghao Sima, Katrin Renz, Kashyap Chitta, Li~Chen, Hanxue Zhang, Chengen Xie,
  Jens Bei{\ss}wenger, Ping Luo, Andreas Geiger, and Hongyang Li.
\newblock {DriveLM: Driving with Graph Visual Question Answering}.
\newblock In \emph{Computer Vision -- ECCV 2024}, pages 256--274. Springer
  Nature Switzerland, 2025.
\newblock \doi{10.1007/978-3-031-72943-0_15}.
\newblock URL
  \url{https://link.springer.com/chapter/10.1007/978-3-031-72943-0_15}.

\bibitem[Tan et~al.(2026)Tan, Yu, Li, Chen, Ju, Luo, Song, and
  Luan]{tan2026msjoe}
Wenhui Tan, Xiaoyi Yu, Jiaze Li, Yijing Chen, Jianzhong Ju, Zhenbo Luo, Ruihua
  Song, and Jian Luan.
\newblock {MSJoE: Jointly Evolving MLLM and Sampler for Efficient Long-Form
  Video Understanding}.
\newblock In \emph{Proceedings of the IEEE/CVF Conference on Computer Vision
  and Pattern Recognition (CVPR)}, pages 19487--19496, 2026.
\newblock URL
  \url{https://openaccess.thecvf.com/content/CVPR2026/html/Tan_MSJoE_Jointly_Evolving_MLLM_and_Sampler_for_Efficient_Long-Form_Video_CVPR_2026_paper.html}.

\bibitem[Tang et~al.(2026)Tang, Han, Sun, Zhou, Zhang, Wei, Yuan, Zhang, Xu,
  and Sun]{tang2026tspo}
Canhui Tang, Zifan Han, Hongbo Sun, Sanping Zhou, Xuchong Zhang, Xin Wei,
  Ye~Yuan, Huayu Zhang, Jinglin Xu, and Hao Sun.
\newblock {TSPO}: Temporal sampling policy optimization for long-form video
  language understanding.
\newblock \emph{Proceedings of the AAAI Conference on Artificial Intelligence},
  40\penalty0 (11):\penalty0 9368--9376, 2026.
\newblock \doi{10.1609/aaai.v40i11.37896}.

\bibitem[Wang et~al.(2026)Wang, Yan, Li, Pan, Li, Zhang, Lu, Xiao, Zhuang, and
  Shen]{wang2026beacon}
Zixuan Wang, Yuchen Yan, Hongxing Li, Teng Pan, Dingming Li, Ruiqing Zhang,
  Weiming Lu, Jun Xiao, Yueting Zhuang, and Yongliang Shen.
\newblock {Milestone-Guided Policy Learning for Long-Horizon Language Agents}.
\newblock In \emph{Proceedings of the 43rd International Conference on Machine
  Learning}, volume 306, pages 131595--131609. PMLR, 2026.
\newblock URL \url{https://proceedings.mlr.press/v306/wang26jm.html}.

\bibitem[Wei et~al.(2026)Wei, Liu, Huang, Peng, Xie, Lin, Jiang, Xu, Yang, Liu,
  Cai, Kang, and Wang]{wei2026evoc2f}
Lei Wei, Qi~Liu, Ruiyang Huang, Xiao Peng, Sicong Xie, Lanbo Lin, Chenhao
  Jiang, Yuanwu Xu, Tianyuan Yang, Jiayao Liu, Li~Cai, Zhaolu Kang, and Bin
  Wang.
\newblock {{E}vo{C}2{F}: Compiling Tool Orchestration for Efficient and
  Evolvable {LLM} Agents}.
\newblock In \emph{Proceedings of the 43rd International Conference on Machine
  Learning}, volume 306, pages 133855--133884. PMLR, 2026.
\newblock URL \url{https://proceedings.mlr.press/v306/wei26q.html}.

\bibitem[Wei et~al.(2025)Wei, Liu, and Ohn-Bar]{wei2025driveqa}
Maolin Wei, Wanzhou Liu, and Eshed Ohn-Bar.
\newblock {Passing the Driving Knowledge Test}.
\newblock In \emph{Proceedings of the IEEE/CVF International Conference on
  Computer Vision (ICCV)}, pages 8395--8406, 2025.
\newblock URL
  \url{https://openaccess.thecvf.com/content/ICCV2025/html/Wei_Passing_the_Driving_Knowledge_Test_ICCV_2025_paper.html}.

\bibitem[Wu et~al.(2026{\natexlab{a}})Wu, Yang, Jiang, Li, Yan, Yu, Zhang,
  Zhai, and Nahrstedt]{wu2025vtoolr1}
Mingyuan Wu, Jingcheng Yang, Jize Jiang, Meitang Li, Kaizhuo Yan, Hanchao Yu,
  Minjia Zhang, ChengXiang Zhai, and Klara Nahrstedt.
\newblock {VTool-R1}: {VLMs} learn to think with images via reinforcement
  learning on multimodal tool use.
\newblock In \emph{International Conference on Learning Representations}, pages
  78298--78319, 2026{\natexlab{a}}.
\newblock URL
  \url{https://proceedings.iclr.cc/paper_files/paper/2026/hash/7e6f445a74cdb71931aac64f1e3f49c9-Abstract-Conference.html}.

\bibitem[Wu et~al.(2026{\natexlab{b}})Wu, Wang, Mei, Cai, Fu, Yang, Wen, Yang,
  Shen, Wang, and Shi]{wu2026evolver}
Rong Wu, Xiaoman Wang, Jianbiao Mei, Pinlong Cai, Daocheng Fu, Cheng Yang,
  Licheng Wen, Xuemeng Yang, Yufan Shen, Yuxin Wang, and Botian Shi.
\newblock {From Interactions to Principles: Experience-Driven Self-Distillation
  for Evolving {LLM} Agents}.
\newblock In \emph{Proceedings of the 43rd International Conference on Machine
  Learning}, volume 306, pages 137024--137044. PMLR, 2026{\natexlab{b}}.
\newblock URL \url{https://proceedings.mlr.press/v306/wu26bf.html}.

\bibitem[Xing et~al.(2025)Xing, Chen, Xie, Xu, Guo, Xu, Hao, Fu, Hu, and
  Heng]{xing2025echotraffic}
Zhenghao Xing, Hao Chen, Binzhu Xie, Jiaqi Xu, Ziyu Guo, Xuemiao Xu, Jianye
  Hao, Chi-Wing Fu, Xiaowei Hu, and Pheng-Ann Heng.
\newblock {EchoTraffic: Enhancing Traffic Anomaly Understanding with
  Audio-Visual Insights}.
\newblock In \emph{Proceedings of the IEEE/CVF Conference on Computer Vision
  and Pattern Recognition (CVPR)}, pages 19098--19108, 2025.
\newblock URL
  \url{https://openaccess.thecvf.com/content/CVPR2025/html/Xing_EchoTraffic_Enhancing_Traffic_Anomaly_Understanding_with_Audio-Visual_Insights_CVPR_2025_paper.html}.

\bibitem[Yao et~al.(2023)Yao, Zhao, Yu, Du, Shafran, Narasimhan, and
  Cao]{yao2023react}
Shunyu Yao, Jeffrey Zhao, Dian Yu, Nan Du, Izhak Shafran, Karthik Narasimhan,
  and Yuan Cao.
\newblock {ReAct}: Synergizing reasoning and acting in language models.
\newblock In \emph{International Conference on Learning Representations
  (ICLR)}, 2023.
\newblock URL \url{https://openreview.net/forum?id=WE_vluYUL-X}.

\bibitem[Yin et~al.(2025)Yin, Lei, and Liu]{yin2025toolvqa}
Shaofeng Yin, Ting Lei, and Yang Liu.
\newblock {ToolVQA: A Dataset for Multi-step Reasoning VQA with External
  Tools}.
\newblock In \emph{Proceedings of the IEEE/CVF International Conference on
  Computer Vision (ICCV)}, pages 4424--4433, 2025.
\newblock URL
  \url{https://openaccess.thecvf.com/content/ICCV2025/html/Yin_ToolVQA_A_Dataset_for_Multi-step_Reasoning_VQA_with_External_Tools_ICCV_2025_paper.html}.

\bibitem[{Z.ai}(2026)]{zai2026glm53flash}
{Z.ai}.
\newblock {GLM-5.3-Flash}: Frontier intelligence, flash cost.
\newblock Z.ai Official Blog, August 2026.
\newblock URL \url{https://z.ai/blog/glm-5.3-flash}.

\bibitem[Zhang et~al.(2026{\natexlab{a}})Zhang, Gu, Li, Ma, Bai, Zhang, Zhang,
  Zhou, He, and Tang]{zhang2026vital}
Haoji Zhang, Xin Gu, Jiawen Li, Chixiang Ma, Sule Bai, Chubin Zhang, Bowen
  Zhang, Zhichao Zhou, Dongliang He, and Yansong Tang.
\newblock {Thinking With Videos: Multimodal Tool-Augmented Reinforcement
  Learning for Long Video Reasoning}.
\newblock In \emph{Proceedings of the IEEE/CVF Conference on Computer Vision
  and Pattern Recognition (CVPR)}, pages 32903--32914, 2026{\natexlab{a}}.
\newblock URL
  \url{https://openaccess.thecvf.com/content/CVPR2026/html/Zhang_Thinking_With_Videos_Multimodal_Tool-Augmented_Reinforcement_Learning_for_Long_Video_CVPR_2026_paper.html}.

\bibitem[Zhang et~al.(2026{\natexlab{b}})Zhang, Chen, Liu, Xue, Liao, Liu,
  Wang, Ning, Chen, Fu, Xie, Sun, Gou, Qi, Meng, Yang, Zhang, Li, Shah, Huynh,
  Li, Yang, Cao, Jang, Zhou, Zhu, Sun, Weston, Su, and
  Wu]{zhang2026earlyexperience}
Kai Zhang, Xiangchao Chen, Bo~Liu, Tianci Xue, Zeyi Liao, Zhihan Liu, Xiyao
  Wang, Yuting Ning, Zhaorun Chen, Xiaohan Fu, Jian Xie, Yuxuan Sun, Boyu Gou,
  Qi~Qi, Zihang Meng, Jianwei Yang, Ning Zhang, Xian Li, Ashish Shah, Dat
  Huynh, Hengduo Li, Zi~Yang, Xuefei Cao, Lawrence~Keunho Jang, Shuyan Zhou,
  Jiacheng Zhu, Huan Sun, Jason~E Weston, Yu~Su, and Yifan Wu.
\newblock {Agent Learning via Early Experience}.
\newblock In \emph{Proceedings of the 43rd International Conference on Machine
  Learning}, volume 306, pages 154443--154475. PMLR, 2026{\natexlab{b}}.
\newblock URL \url{https://proceedings.mlr.press/v306/zhang26m.html}.

\bibitem[Zhang et~al.(2026{\natexlab{c}})Zhang, Zhao, Luo, Li, and
  Tang]{zhang2026trafficvqa}
Yu~Zhang, Zhicheng Zhao, Ze~Luo, Chenglong Li, and Jin Tang.
\newblock {UAV traffic scene understanding: A regulation embedded multi-modal
  network and a unified benchmark}.
\newblock \emph{arXiv preprint arXiv:2603.10722}, 2026{\natexlab{c}}.
\newblock URL \url{https://arxiv.org/abs/2603.10722}.

\bibitem[Zhang et~al.(2026{\natexlab{d}})Zhang, Fang, Chen, and
  Pechenizkiy]{zhang2026qevolve}
Yudi Zhang, Meng Fang, Zhenfang Chen, and Mykola Pechenizkiy.
\newblock {Self-evolving LLM agents with in-distribution Optimization}.
\newblock In \emph{Proceedings of the 43rd International Conference on Machine
  Learning}, volume 306, pages 155348--155365. PMLR, 2026{\natexlab{d}}.
\newblock URL \url{https://proceedings.mlr.press/v306/zhang26au.html}.

\bibitem[Zhao et~al.(2024)Zhao, Huang, Xu, Lin, Liu, and Huang]{zhao2024expel}
Andrew Zhao, Daniel Huang, Quentin Xu, Matthieu Lin, Yong-Jin Liu, and Gao
  Huang.
\newblock {ExpeL: LLM Agents Are Experiential Learners}.
\newblock In \emph{Proceedings of the AAAI Conference on Artificial
  Intelligence}, volume~38, pages 19632--19642. {AAAI} Press, 2024.
\newblock \doi{10.1609/aaai.v38i17.29936}.
\newblock URL \url{https://ojs.aaai.org/index.php/AAAI/article/view/29936}.

\bibitem[Zheng et~al.(2025)Zheng, Liu, Li, Chen, Yu, Gao, Dang, Liu, Men, Yang,
  Zhou, and Lin]{zheng2025gspo}
Chujie Zheng, Shixuan Liu, Mingze Li, Xiong-Hui Chen, Bowen Yu, Chang Gao, Kai
  Dang, Yuqiong Liu, Rui Men, An~Yang, Jingren Zhou, and Junyang Lin.
\newblock {Group Sequence Policy Optimization}.
\newblock \emph{arXiv preprint arXiv:2507.18071}, 2025.
\newblock URL \url{https://arxiv.org/abs/2507.18071}.

\bibitem[Zheng et~al.(2026)Zheng, Yang, Hong, Zhao, Xu, Yang, Shen, and
  Yu]{zheng2026deepeyes}
Ziwei Zheng, Michael Yang, Jack Hong, Chenxiao Zhao, Guohai Xu, Le~Yang, Chao
  Shen, and Xing Yu.
\newblock {DeepEyes}: Incentivizing ``thinking with images'' via reinforcement
  learning.
\newblock In \emph{International Conference on Learning Representations}, pages
  126775--126798, 2026.
\newblock URL
  \url{https://proceedings.iclr.cc/paper_files/paper/2026/hash/cdb347d7516d52a9280ea9d6708911f6-Abstract-Conference.html}.

\bibitem[Zhou et~al.(2025)Zhou, Larintzakis, Guo, Zimmer, Liu, Cao, Zhang,
  Lakshminarasimhan, Strand, and Knoll]{zhou2025tumtraf}
Xingcheng Zhou, Konstantinos Larintzakis, Hao Guo, Walter Zimmer, Mingyu Liu,
  Hu~Cao, Jiajie Zhang, Venkatnarayanan Lakshminarasimhan, Leah Strand, and
  Alois Knoll.
\newblock {{TUMT}raf {V}ideo{QA}: Dataset and Benchmark for Unified
  Spatio-Temporal Video Understanding in Traffic Scenes}.
\newblock In \emph{Proceedings of the 42nd International Conference on Machine
  Learning}, volume 267, pages 78786--78811. PMLR, 2025.
\newblock URL \url{https://proceedings.mlr.press/v267/zhou25g.html}.

\end{thebibliography}

\clearpage
\appendix
\section*{Supplementary Material}
\section{Evaluation on Real-World Traffic Dataset}
\label{sec:private-client}

\begin{table*}[!htbp]
\centering
\caption{Performance comparison on a private
traffic-violation review dataset.
We report accuracy (Acc.) and F1 for each category.
Overall accuracy is the mean of the per-group accuracies.
Overall F1 is the sample-weighted mean across the six
categories. Bold and underline indicate the best
and second-best results, respectively.}
\label{tab:private-client}

\scriptsize
\setlength{\tabcolsep}{2pt}
\renewcommand{\arraystretch}{1.1}

\resizebox{\textwidth}{!}{%
\begin{tabular}{@{}l*{14}{c}@{}}
\toprule
\textbf{Method}
& \multicolumn{2}{c}{\textbf{Seatbelt}}
& \multicolumn{2}{c}{\textbf{Red light}}
& \multicolumn{2}{c}{\textbf{Phone use}}
& \multicolumn{2}{c}{\textbf{Solid line}}
& \multicolumn{2}{c}{\textbf{Lane use}}
& \multicolumn{2}{c}{\textbf{Wrong-way}}
& \multicolumn{2}{c}{\textbf{Overall}} \\
\cmidrule(lr){2-3}
\cmidrule(lr){4-5}
\cmidrule(lr){6-7}
\cmidrule(lr){8-9}
\cmidrule(lr){10-11}
\cmidrule(lr){12-13}
\cmidrule(lr){14-15}

& Acc. & F1
& Acc. & F1
& Acc. & F1
& Acc. & F1
& Acc. & F1
& Acc. & F1
& Acc. & F1 \\
\midrule

Qwen3.8-Flash-Next
& \underline{0.5226} & 0.6631
& 0.5179 & 0.6519
& 0.7337 & 0.7337
& 0.5464 & 0.6573
& 0.5484 & 0.5385
& \underline{0.8108} & 0.7835
& 0.6107 & 0.6670 \\

Qwen3.8-27B
& 0.5155 & \underline{0.6667}
& 0.5298 & 0.6728
& \textbf{0.8191} & \textbf{0.8468}
& 0.5269 & 0.6271
& 0.5848 & 0.6226
& 0.7818 & 0.7209
& 0.6166 & 0.6793 \\

DeepSeek-V4.1-Flash
& 0.4949 & 0.4762
& 0.5235 & 0.6816
& 0.5567 & 0.3857
& \underline{0.5707} & \underline{0.6867}
& 0.5368 & 0.5749
& 0.5852 & 0.3348
& 0.5455 & 0.5213 \\

GLM-5.3-Flash
& 0.5170 & 0.5824
& \underline{0.6250} & \underline{0.7179}
& 0.4796 & 0.5952
& 0.5194 & 0.6601
& 0.4791 & 0.6383
& 0.7009 & 0.5455
& 0.5538 & 0.6166 \\

Qwen3-VL-32B
& 0.5025 & 0.6633
& 0.5900 & 0.6985
& 0.5779 & 0.7021
& 0.5500 & 0.6751
& \textbf{0.8350} & \textbf{0.8514}
& 0.8075 & \underline{0.8300}
& \underline{0.6558} & \underline{0.7440} \\

\midrule

\textbf{StarTV-Agent}
& \textbf{0.7188} & \textbf{0.7682}
& \textbf{0.7663} & \textbf{0.7981}
& \underline{0.7789} & \underline{0.8167}
& \textbf{0.6788} & \textbf{0.7220}
& \underline{0.8135} & \underline{0.8085}
& \textbf{0.8112} & \textbf{0.8419}
& \textbf{0.7590} & \textbf{0.7896} \\

\bottomrule
\end{tabular}%
}
\end{table*}

\noindent\textbf{Data and Setup.}
We further evaluate StarTV-Agent on a private
traffic-violation review dataset collected from a
municipal traffic-police deployment.
The dataset contains 1,999 binary VQA samples
across six categories: seatbelt violations,
red-light violations, phone use while driving, 
solid-line crossing, improper lane use, and wrong-way driving.
These recurring review tasks provide a practical
setting for evaluating StarTV-Agent on real-world
traffic data. We compare against five VLM
baselines: Qwen3.8-Flash-Next~\cite{qwen2026qwen38flashnext},
Qwen3.8-27B~\cite{qwen2026qwen38},
DeepSeek-V4.1-Flash~\cite{deepseek2026v41flash},
GLM-5.3-Flash~\cite{zai2026glm53flash}, and
Qwen3-VL-32B~\cite{bai2025qwen3vl}.
We report accuracy and F1 in
Table~\ref{tab:private-client}.
The data were collected and used with proper authorization
but cannot be publicly released due to privacy and
data-sharing restrictions.

\noindent\textbf{Results.}
As shown in Table~\ref{tab:private-client},
StarTV-Agent achieves the highest overall
accuracy and F1 score,
outperforming the strongest VLM baseline,
Qwen3-VL-32B, by 15.7\% and 6.1\%, respectively, in relative terms.
It ranks first in both metrics on four of the
six violation categories, with particularly
large accuracy gains for seatbelt violations and red-light violations.
Qwen3.8-27B performs best on phone-use detection,
while Qwen3-VL-32B leads on improper lane use.
These results extend the evaluation of StarTV-Agent
to real-world traffic-violation review and show
stronger overall performance than VLM baselines.

\section{Datasets and Evaluation Protocols}
\label{app:datasets}

\noindent\textbf{TUMTraf VideoQA.}
TUMTraf VideoQA~\cite{zhou2025tumtraf} evaluates
question answering on roadside traffic videos,
requiring spatial and temporal reasoning.
It covers different categories---positioning, counting,
motion, class, and existence---assessing capabilities
ranging from object localization and recognition to
motion understanding and presence reasoning.
All methods receive the same 16 sampled frames per video.

\noindent\textbf{Traffic-VQA.}
Traffic-VQA~\cite{zhang2026trafficvqa} evaluates traffic understanding from UAV imagery using optical (OPT) and thermal infrared (TIR) observations. The two sensing modalities provide complementary visual information but differ substantially in appearance and object representation, posing challenges for robust perception across imaging conditions and effective multimodal reasoning. We evaluate the same questions under three input settings: OPT, TIR, and multimodal (MUL). For MUL, optical and thermal observations are provided as two ordered, separately identified images, requiring models to integrate complementary evidence across modalities. We report accuracy for each setting and their arithmetic mean.

\noindent\textbf{RoadSceneVQA.}
RoadSceneVQA~\cite{guan2026roadscenevqa} evaluates open-ended question answering on roadside imagery, requiring models to interpret complex traffic scenes and generate answers grounded in visual evidence. Unlike multiple-choice evaluation, its open-ended setting requires models to identify relevant scene elements, understand their attributes and relationships, and formulate accurate responses without relying on predefined answer options. This setting therefore provides a complementary assessment of fine-grained visual understanding and answer generation. We evaluate all methods on its test set using the seven metrics in Table~\ref{tab:roadscene_results}. For GPT-Score, we use GPT-5.5~\cite{openai2026gpt55} as the judge, with identical prompts and request settings across methods.

\section{Adaptation of Tool-Using Baselines}
\label{app:baselines}

We adapt DeepEyes~\cite{zheng2026deepeyes}, VTool-R1~\cite{wu2025vtoolr1}, and TAU-Agent~\cite{lin2026tauagent} to a unified evaluation protocol. All agents receive the same prepared inputs: 16 temporally ordered frames with timestamps for TUMTraf VideoQA, and a single OPT/TIR image or an ordered OPT--TIR pair for Traffic-VQA. Paired modalities are identified as observations of the same scene, and no agent can access additional video frames. Predictions follow each benchmark's native answer format and scoring rules.

\noindent\textbf{DeepEyes.}
We preserve its zoom-in operation by mapping the requested bounding box and one-based image index to a crop returned in the next reasoning turn.

\noindent\textbf{VTool-R1.}
We replace its original visual-editing operations with the \texttt{crop\_image()} interface over the supplied images, without executing arbitrary Python code.

\noindent\textbf{TAU-Agent.}
We retain its captioning, tracking, retrieval,
and question-answering stages, adapting them
to fixed frame bundles instead of the original
video input. We use Qwen3.8-Max~\cite{alibaba2026qwen38max}
for its subtitle generation and retrieval.

\section{Training Data and Reward Design}
\label{app:training_data}

\paragraph{Training Data Preparation.}
We build a unified training set from the official training
annotations of TUMTraf VideoQA~\cite{zhou2025tumtraf},
Traffic-VQA~\cite{zhang2026trafficvqa}, and
RoadSceneVQA~\cite{guan2026roadscenevqa}.
The original questions, answers, media identifiers,
and response formats are preserved.
After removing empty and duplicate entries,
we reserve approximately 5\% of the training media
from each benchmark for validation (seed 42).
Questions sharing the same image or video are kept
in the same split, and media overlapping with the
official test sets are excluded.
We then sample up to 2,000 questions balanced
across the three benchmarks for RL training.
Videos are represented by 16 ordered frames with
timestamps, while optical--thermal pairs remain
separate and ordered.
Reference answers are used only for training-time
evaluation and are never provided to the Student policy.

\paragraph{Terminal Task Reward.}
The terminal reward is computed from the final answer
according to the task requirements.
A rollout is scored only when execution completes
successfully and its evidence sources are valid.
Otherwise, answer extraction returns an empty string
and the task reward is set to zero.
Table~\ref{tab:terminal_reward} summarizes the
scoring rules for different answer types.
Semantic scores for open-ended tasks serve as
training rewards rather than official benchmark metrics.

\begin{table}[t]
\centering
\caption{Terminal task reward rules used by the
unified evaluator.}
\label{tab:terminal_reward}

\footnotesize
\setlength{\tabcolsep}{4pt}
\renewcommand{\arraystretch}{1.15}

\begin{tabularx}{\columnwidth}{
@{}>{\raggedright\arraybackslash}p{0.25\columnwidth}
>{\raggedright\arraybackslash}X@{}}
\toprule
\textbf{Answer Type} & \textbf{Reward and Scoring Rule} \\
\midrule

Binary, choice,
and short VQA
& Exact match ($1$ if correct, $0$ otherwise).
Task-aware matching avoids substring checks
for binary answers. \\

\addlinespace[2pt]

Grounding
& $r_{\mathrm{task}}=1-\min(1,E_{\mathrm{ST}})$,
where $E_{\mathrm{ST}}$ is the spatio-temporal
grounding error. The original error is also
recorded for analysis. \\

\addlinespace[2pt]

Open VQA
and caption
& Normalized exact matches receive $1$;
otherwise, a semantic judge assigns a score
in $[0,1]$ based on the reference answer. \\

\bottomrule
\end{tabularx}
\end{table}

\paragraph{Role-Specific Process Scoring.}
For RSPO, an independent Process Teacher scores
each Student decision based on its role, available
evidence, and execution feedback.
The score $p_{i,t}\in[0,1]$ reflects the quality of
the decision already made, rather than a suggested
alternative. Training references are available to
the Process Teacher but never to the Student.
Table~\ref{tab:process_reward} provides the shared
instruction and role-specific scoring prompts.
Only responses containing a valid numeric
\texttt{score} in $[0,1]$ are accepted.

\begin{table}[t]
\centering
\footnotesize
\setlength{\tabcolsep}{3pt}
\renewcommand{\arraystretch}{1.12}
\caption{Shared instruction and role-specific scoring
prompts used by the Process Teacher.}
\label{tab:process_reward}
\begin{tabular}{
@{}p{0.20\columnwidth}
p{0.75\columnwidth}@{}}
\toprule
\textbf{Scope} & \textbf{Prompt} \\
\midrule

Shared
& ``Judge this one already-made Student decision.
Score its decision quality in [0,1].
Return only JSON with one numeric key named score.
Do not suggest a new action. Assess action legality,
including validation\_error, as well as task relevance.'' \\

\midrule

Planner
& ``Is the plan relevant to the question,
grounded in visible input, and executable
with available tools?'' \\

Tool Selector
& ``Is the selected next tool and subgoal
appropriate for the current evidence
and available tools?'' \\

Executor
& ``Are parameters and resource references
legal, available now, and suitable for
the selected tool?'' \\

Verifier
& ``Does the continue/stop decision faithfully
reflect actual tool output and evidence?'' \\

Summary
& ``Is the answer supported by observed evidence,
does it answer the user question and follow
the public task contract?'' \\

\bottomrule
\end{tabular}
\end{table}

\section{Tool Inventory}
\label{app:tool_inventory}

StarTV-Agent integrates 22 tools across four categories.
\emph{Visual reasoning and attribute extraction}
supports image understanding, attribute recognition,
and visual annotation.
\emph{Object and road-scene localization}
identifies traffic participants and road elements.
\emph{Traffic-specific assessment}
detects traffic behaviors and violations.
\emph{Evidence acquisition and aggregation}
supports frame extraction, retrieval, and object counting.
As summarized in Table~\ref{tab:tool_inventory},
each tool defines its required inputs and returned evidence.
All learned expert models remain frozen.
Typed resource references and source dependencies
enable evidence integration across reasoning steps.

\begin{table*}[t]
\centering
\caption{Tool inventory of StarTV-Agent, grouped by functionality.
Inputs specify the required media, prompts, or resources,
while outputs describe the evidence returned for subsequent reasoning.}
\label{tab:tool_inventory}

\begingroup
\footnotesize
\setlength{\tabcolsep}{3pt}
\renewcommand{\arraystretch}{1.12}

\begin{tabular}{@{}p{0.34\textwidth} p{0.25\textwidth} p{0.35\textwidth}@{}}
\hline
\textbf{Tool} & \textbf{Required Input} & \textbf{Returned Evidence} \\
\hline

\multicolumn{3}{@{}l}{\emph{Visual reasoning and attribute extraction}} \\

\texttt{General\_Generator\_Tool}
& Image resources and a prompt
& Free-form, image-grounded descriptions or reasoning results. \\

\texttt{Multi\_Image\_VQA\_Tool}
& Ordered images and a question
& Joint visual reasoning results, including temporal and cross-image comparisons. \\

\texttt{VLM\_Attribute\_Extraction\_Tool}
& Images or object crops and target attributes
& Structured attribute--value pairs for each image. \\

\texttt{general\_ocr}
& Image
& Text blocks with content, confidence scores, and polygon coordinates. \\

\texttt{Draw\_Tool}
& Image and geometric primitives (box/point/line/polygon)
& Annotated image with drawn overlays. \\

\hline
\multicolumn{3}{@{}l}{\emph{Object and road-scene localization}} \\

\texttt{YOLO\_Detector\_Tool}
& Image and predefined object classes
& Object labels, bounding boxes, and confidence scores. \\

\texttt{OWD\_Object\_Detector\_Tool}
& Image and open-vocabulary text prompts
& Prompt-matched labels, bounding boxes, and confidence scores. \\

\texttt{SAM3\_Tool}
& Image and text prompts
& Prompt-matched labels, bounding boxes, and confidence scores. \\

\texttt{Road\_Mark\_Object\_Detector\_Tool}
& Image and supported road-marking classes
& Road-marking labels, bounding boxes, polygons, and confidence scores. \\

\texttt{Lane\_Arrow\_Direction\_Detect}
& Image
& Lane-arrow directions, bounding boxes, and confidence scores. \\

\texttt{illegal\_parking}
& Image
& Motor/non-motor vehicle boxes, fine-grained vehicle types
(car/SUV/bus/truck/van, etc.), and attributes. \\

\texttt{fpvn\_tool}
& Image
& Face, person, vehicle, and non-vehicle detections
with bounding boxes and confidence scores. \\

\texttt{human\_detection}
& Image
& Head and body bounding boxes with confidence scores. \\

\hline
\multicolumn{3}{@{}l}{\emph{Traffic-specific assessment}} \\

\texttt{RedLight\_Violation\_Detector\_Tool}
& Three ordered observations and a target-vehicle box
& Red-light violation assessment for the target vehicle. \\

\texttt{Phone\_Classifier\_Tool}
& Cropped person image
& Driving-related phone-use classification. \\

\texttt{Seatbelt\_Classifier\_Tool}
& Cropped person image
& Seatbelt-use classification. \\

\texttt{rider\_no\_helmet}
& Image
& Non-helmeted rider boxes and riding types
(e-bike/motorcycle/tricycle, etc.). \\

\texttt{nv\_overload\_detection}
& Image
& Overloaded non-motor vehicles with bounding boxes,
vehicle types, and rider counts. \\

\hline
\multicolumn{3}{@{}l}{\emph{Evidence acquisition and aggregation}} \\

\texttt{Video\_Frame\_Tool}
& Video resource and requested timestamp
& Extracted image resource and actual frame timestamp. \\

\texttt{Search\_Tool}
& Text retrieval query
& Retrieved images and references from the configured database. \\

\texttt{Resource\_Count\_Tool}
& Object resources and counting rules
& Matched objects, rule-specific counts, and aggregate counts. \\

\texttt{human\_crowd\_piling}
& Image
& Head bounding boxes with \texttt{group\_id} clusters
and confidence scores. \\

\hline
\end{tabular}
\endgroup
\end{table*}

\FloatBarrier
\section{Prompt Design}

\label{app:training_prompts}

\subsection{Role-Specific Decision Prompts}

\subsubsection{Planner}

\begin{stvprompt}{Planner / System instruction}
You are a Routine workflow Complex Visual Task Planner. Your role is to analyze the user query, image content, and available tools to generate a structured execution plan.

- Select tools by their actual capability, evidence requirements and sufficient output quality, then compare workflow cost using the shared tool selection policy.
- For object categories, use a fixed-class detector only if its advertised label_options contains the exact requested semantic label. A broader category is not enough. For open-domain labels, use an open-vocabulary detector on IMAGE resources.
- Every tool in the "tools" list must exist in Toolbox_metadata (exact name match).
- The plan must be logically coherent: each step's output feeds the next step naturally.

- "ismatch": Whether the image scene matches the user query. This is a strict gate -- returning "NO" will terminate the task. Only return "NO" when the image clearly conflicts with the query intent. When in doubt, return "YES".
\end{stvprompt}

\begin{stvprompt}{Planner / Dynamic user input}
Toolbox_metadata:{{ toolbox_metadata }}
Template:{{ template }}

1. Capability-based tool selection:
- Refer to the tool list **Toolbox_metadata**.
- Check actual capability, input evidence and required outputs in the tool metadata.
- Select an applicable, sufficiently effective tool or decompose the task; apply the shared tool selection policy.

6. Select tools by their advertised capabilities. Use only exact names from
Toolbox_metadata. If no available tool can perform the task, return empty
tools and execution_flow lists and explain the limitation in thought.
7. Return one JSON object with exactly the fields below, without Markdown or
additional fields. subtasks, tools and execution_flow are arrays of strings.
ismatch is YES when the image is consistent with the query, NO only when it
clearly conflicts, and UNKNOWN when no image is provided.
{{ native_action_example("planner") }}
\end{stvprompt}

\begin{stvprompt}{Planner / Native action example}
{
  "thought": "Brief reasoning grounded in the available evidence.",
  "ismatch": "YES",
  "subtasks": [
    "Describe the next subtask."
  ],
  "tools": [
    "Available_Tool_Name"
  ],
  "execution_flow": [
    "Run the available tool for the subtask."
  ]
}
\end{stvprompt}

\subsubsection{Tool Selector}

\begin{stvprompt}{Tool Selector / System instruction}
You are a visual workflow task planner. Select the NEXT tool based on the Query Analysis plan AND actual execution feedback.
Decision Logic:
- Query Analysis is a directional anchor, not a rigid script.
- Execution results are reality. They may invalidate parts of the plan.
- Your job: bridge plan intent with execution reality.
Core Rules:
1. Read Query Analysis -> find the provisional next sub-task.
2. Check Execution History -> does reality support it?
   - Already satisfied by prior output -> skip, move to next sub-task.
   - Prior tool failed/empty -> pivot (different tool/params/approach).
   - Unexpected finding -> adjust sub-goal, keep direction.
   - Nothing contradicts -> follow the plan.
3. Select ONE tool for the dynamically-determined next sub-goal.
4. Never repeat a tool+goal+input combination already attempted.
\end{stvprompt}

\begin{stvprompt}{Tool Selector / Dynamic user input}
Query: {{ query_input }}
Query Analysis: {{ planner }}
Available Tools: {{ available_tools }}
Tool Metadata: {{ available_toolbox_metedata }}
**Available Resources** (already detected/processed, referenced by ID):
{{ resource_summary }}
**Execution History** (completed steps, DO NOT repeat same tool+goal):
{{ memory }}

{% if has_verifier_suggestion %}
## Verifier Feedback (from previous step evaluation)
The Verifier reviewed the last step and determined it was insufficient. Pay close attention to this guidance:
{{ verifier_suggestion }}

Anti-Repetition:
- Same tool + same goal + same input -> FORBIDDEN.
- Tool returned empty -> retry ONCE with a different tool of the same category.
- Same tool on different input (e.g., crop/zoom) -> NOT repetition, it's progressive analysis.
- No non-repeating step available -> output "No suitable next step" in thought.
\end{stvprompt}

\begin{stvprompt}{Tool Selector / Native action example}
{
  "thought": "Brief reasoning grounded in the available evidence.",
  "context": "Relevant existing evidence and resource IDs.",
  "sub_goal": "The next subtask.",
  "tool_name": "Available_Tool_Name"
}
\end{stvprompt}

\subsubsection{Executor}

\begin{stvprompt}{Executor / System instruction}
You are a tool parameter generator. Your output has TWO categories:

## 1. Framework Parameter: `resource_filters` (MUST generate)
How to reference existing resources. A LIST of filter dicts; set to `null` to work on the original input image.

Each filter dict supports: type, label, label_in, attr, attr_in, attr_filter,
min_confidence, source_tool, source_tool_in, tags, has_attr, max_results.

## 2. Business Parameters: `params` (MUST generate)
Tool-specific parameters matching the tool's `input_types` schema.

## 3. System Parameters (NEVER generate -- framework resolves from resource_filters)
These are automatically injected by the framework. Do NOT put them in `params` or anywhere:
\end{stvprompt}

\begin{stvprompt}{Executor / Resource binding and tool schema}
Query: {{ query_input }}
Context: {{ context }}
Sub-Goal: {{ sub_goal }}
Selected Tool: {{ tool_name }}
Tool Metadata: {{ tool_metadata }}

You MUST generate TWO fields: `resource_filters` (framework parameter) and `params` (business parameters).
NEVER generate system parameters: image, image_path, objects, bbox, coordinates, image_resource_ids, object_resource_ids, etc. -- the framework resolves these from resource_filters automatically.

- Supply every required_params field. Match the declared scalar/list type exactly; a float time_seconds is one timestamp, not a list.
- For finite-class detectors, choose labels only from user_metadata.label_options. Instance identifiers are not semantic labels.
- For image-only tools, select IMAGE or detected OBJECT resources. Extract VIDEO frames before using image tools.

- Fields within one filter entry = AND
- _in fields within a list = OR
- Multiple filter entries = independent resource groups (merged by type)
\end{stvprompt}

\begin{stvprompt}{Executor / Parameter binding example}
{
  "resource_filters": [
    {"type": "object", "label": "person",
     "min_confidence": 0.8}
  ],
  "params": {"attributes": ["gender", "age", "upper_color"]}
}
\end{stvprompt}

\subsubsection{Verifier}

\begin{stvprompt}{Verifier / System instruction}
You are a task completeness verifier. Your job is to determine whether the accumulated execution results are sufficient to answer the user's query.
You have three responsibilities:
1. Judge: STOP (sufficient evidence) or CONTINUE (need more steps).
2. Guide: When CONTINUE, provide a concrete suggestion to help the Tool Selector choose the next step correctly.
3. Filter: When STOP, specify result_filters to select the final answer resources from the accumulated results.
You do NOT:
- Produce final natural-language answers
You ONLY judge, guide, and filter.
Execution evidence rules:
- A tool's successful status proves only that it ran. A failed resource filter proves only that no resource matched that filter; it does not prove the target is absent.
- Draw_Tool output is a visualization of upstream geometry. It does not establish object identity, tracking, motion, or other visual properties. Require the relevant analysis tool's result for those claims.
- A connection error or HTTP 5xx is a service failure, not a negative detection result. Do not produce a definite answer from that failed call.
\end{stvprompt}

\begin{stvprompt}{Verifier / Evidence validation and decision}
Query: {{ query_input }}
Query Analysis: {{ planner }}
Sub Goal: {{ sub_goal }}
Execution History: {{ memory }}
**Available Resources**:
{{ resource_summary }}

- A requirement is fulfilled only when a tool-produced result covers it -- you must NOT treat "you could derive it from existing resources" as fulfilled
- A rendered or annotated image only shows geometry from its parent object resources. It cannot itself satisfy detection, tracking, or state requirements. Motion or stationary claims require evidence from multiple frames with target association and motion analysis.
- A failed filter, failed tool call, or HTTP error is not an empty detection. Only an explicit successful detector result with zero detections can support target absence.

Step 4 -- Decide
- STOP: all requirements map to concrete, tool-produced, visually-consistent resources
- CONTINUE: any requirement lacks tool-produced resources, OR resources conflict with visual evidence

SUGGESTIONS (when CONTINUE):
- Empty last step -> describe what's missing and suggest alternative approaches (e.g., broader search, lower threshold)
- Missing attributes -> specify what resources need which attributes
- Failed step -> identify likely cause and alternative strategy
- Visual inconsistency -> suggest re-examination or adjusted strategy; do not accept questionable results
- Describe what needs to be done next, not which tool to use
\end{stvprompt}

\begin{stvprompt}{Verifier / Resource-filter semantics}
Core Principle: The role of result_filters is to select resources. First classify the Query into info/condition type, then construct filters according to the corresponding rules. result_filters must derive from actual Available Resources, not query text alone.

- Info type: The answer comes from existing attribute values of the target object; filters must NOT pre-judge the answer
- "Is there / Does / What / Where / How many" -> The answer is determined by attribute values; filters only select the target object
- Condition type: The answer is a set of objects meeting specific criteria; the criteria values ARE the filters
- "Find / Retrieve / Mark / Box + criteria" -> The answer is a set of matching objects; filters precisely specify the criteria
- Composite type: Contains both info and condition sub-questions -> Each sub-question constructs an independent filter group per its own rules

null = accept all resources (only when the query is a description type, e.g., "describe the image")
\end{stvprompt}

\begin{stvprompt}{Verifier / Native action example}
{
  "thought": "Brief reasoning grounded in the available evidence.",
  "Conclusion": "CONTINUE",
  "Explain": "Evidence supporting this decision.",
  "result_filters": null,
  "suggestion": "The next evidence needed."
}
\end{stvprompt}

\subsubsection{Summary}

\begin{stvprompt}{Summary / System instruction}
You are a task result summarizer. Given a user query and a set of detected/analyzed resources, provide a concise summary that directly answers the query.
RESPONSE FORMAT (strictly follow this JSON structure without any additional text, explanations, or markdown formatting):
{
  "Explain": "str, Brief explanation in Chinese (max 100 words) - reasoning based on the provided resources",
  "Answer": "str, One-sentence final answer to the query"
}
IMPORTANT: Output ONLY the valid JSON structure above without any markdown backticks, comments, or additional explanatory text. Do not include any prefix or suffix text.
\end{stvprompt}

\begin{stvprompt}{Summary / Dynamic user input}
Query: {{ query_input }}
**Available Resources** (results from execution):
{{ resource_summary }}
Based on the Available Resources, summarize the findings and answer the query. Do not fabricate information not present in the resources.
\end{stvprompt}

\subsection{Shared Tool-Selection Instruction}

\begin{stvprompt}{Shared selection policy / System text}
Tool selection policy (applies to plans, templates and runtime choices):
1. Capability and evidence come first. Check input modality, exact supported labels,
   use_scenario/support/not_support/limitation, required resources and output types.
   Never substitute a broader label or an unsupported tool to save cost. A template
   cannot override a tool's capability or evidence requirements.
2. Judge whether the expected output is sufficient for this subtask using actual
   observations and failures. user_metadata.performance_coefficient is a rough,
   in-scope effectiveness PRIOR, not measured accuracy, detection confidence or a
   probability of success. Compare it only for tools that can satisfy the same need.
   Model priors may favor more capable, expensive services, but do not guarantee that
   they outperform a specialist on its own task. Deterministic tools' 1.0 describes
   the operation on valid inputs, not the completeness of upstream evidence.
3. Among sufficiently effective, applicable options, reuse existing evidence and
   compare the expected TOTAL workflow cost. user_metadata.cost_coefficient is
   relative resource cost per typical logical call (YOLO single-image call = 1.0),
   not measured latency or API price. Image count/resolution, video seeking and
   generation length change actual cost. Missing coefficients mean unknown.
4. Use an applicable higher-cost tool directly when cheaper tools lack the needed
   capability; upgrade when their output is insufficient, unreliable or has failed.
   Do not force a cheap failed attempt, a fixed tool chain, or performance/cost-ratio
   ranking. Specialized tools are not unconditionally preferred.
5. An empty detection or resource-filter miss is not proof of absence or a complete
   count. Free-text VQA does not supply verified geometry or temporal evidence.
   If evidence is unavailable, report the gap instead of inventing it.
These coefficients are read-only metadata. Do not emit them in actions, params,
resource_filters, HTTP arguments or final answers. Use only the native action schema.
\end{stvprompt}

\subsection{Training-Time Scoring Prompts}

\begin{stvprompt}{Process Teacher / System scaffold}
Judge this one already-made Student decision. Score its decision quality in [0,1]. Return only JSON with one numeric key named score. Do not suggest a new action. Assess action legality, including validation_error, as well as task relevance. Role-specific rubric: {{role_rubric}} The training-only reference is for judging; it must never enter a Student request. {"ground_truth": {{reference_answer}}, "reference_evidence_ids": {{reference_evidence_ids}}}
\end{stvprompt}

\begin{stvprompt}{Process Teacher / Dynamic request fields}
{"event_id": "{{event_id}}",
 "decision_type": "{{role}}",
 "task_contract": {{task_contract}},
 "student_messages": {{student_messages_with_image_payloads_redacted}},
 "student_action_raw": {{raw_response}},
 "student_action_parsed": {{parsed_response}},
 "validation_error": {{validation_error}},
 "repair_index": {{repair_index}},
 "available_tools": {{available_tools}},
 "selected_tool": {{selected_tool}},
 "parameter_names": {{parameter_names}}}
[Original image/frame payloads appended as image_url parts.]
\end{stvprompt}

\begin{stvprompt}{Terminal semantic judge / System message}
Score semantic equivalence and factual completeness of the prediction relative to the reference. Return exactly one JSON object with a numeric key named score, for example {"score": 0.5}. The score must be in [0,1]; contradictions score 0. Do not include explanations or any other keys.
\end{stvprompt}

\begin{stvprompt}{Terminal semantic judge / User message}
Prediction: {{prediction}}
Reference: {{reference}}
\end{stvprompt}

\subsection{Workflow Experience Evolution Prompts}

\subsubsection{Shared Memory-Maintenance Constraints}

\begin{stvprompt}{Workflow maintenance / Shared system rules}
Maintain a small library of reusable procedures from recent REAL executions.
Observed answers may be WRONG: never memorize an answer, count, label or source value.
Do not infer correctness from completion status.

For a TEMPLATE, content MUST be an object with contiguous step_1, step_2, ... keys; each step is {tool_name,sub_goal,parameters}.
tool_name MUST be spelled EXACTLY as in the supplied tools list -- never invent, translate or wrap a tool name.
parameters MUST be {"params":{...}} optionally plus "resource_filters":[...]; no other top-level key is allowed.
Every key inside params MUST be one of that tool's legal_params; do not use placeholder names such as unknown_argument or unknown_parameter.
Use only generic constants or resource_filters, never a question-specific value, path, geometry or resolved ID.
No slots.
Describe procedures, never classification code or a hard tool whitelist.
\end{stvprompt}

\subsubsection{Workflow Addition}

\begin{stvprompt}{Addition / Branch-specific system instruction}
Add at most ONE new experience with parent_id:null.
Return {"action":"noop"} when nothing is worth keeping; otherwise
{"action":"add","candidate":{"parent_id":null,
 "kind":"template","operation":string,"question":string,
 "change":"content","requirements":{"modalities":[],
 "temporal":false,"min_images":0,"notes":string,
 "reference_frame":"","reference_mode":""},
 "content":object,"slots":{}}}.
\end{stvprompt}

\subsubsection{Workflow Revision}

\begin{stvprompt}{Revision / Branch-specific system instruction}
Revise ONE of <target-id-list> on its SAME id (parent_id = that id).
For a revision set action to "revise", put the exact target id in parent_id, reuse that target's kind, and set change to "content" or "boundary".
A revision MUST keep parent_id equal to a supplied revision target id.

If contract_feedback is non-empty, your previous template was REJECTED for that reason: fix exactly that field before proposing again.
counterexamples list real tool steps taken when a revision target was hit: use them to correct the procedure, never to copy an answer.
A boundary revision changes ONLY requirements: omit content and the parent's flow is kept; a content revision keeps the parent's requirements unless they are the actual problem.
\end{stvprompt}

\end{document}